\documentclass{article} 
\usepackage{iclr2022_conference,times}

\iclrfinalcopy

\usepackage{amsmath,amsfonts,bm}

\def\eqref#1{equation~\ref{#1}}

\def\1{\bm{1}}

\def\vc{{\bm{c}}}
\def\vd{{\bm{d}}}
\def\ve{{\bm{e}}}

\def\vh{{\bm{h}}}

\def\vl{{\bm{l}}}

\def\vp{{\bm{p}}}

\def\vs{{\bm{s}}}
\def\vt{{\bm{t}}}

\def\vx{{\bm{x}}}

\def\vz{{\bm{z}}}

\def\mA{{\bm{A}}}

\def\mI{{\bm{I}}}

\def\mL{{\bm{L}}}
\def\mM{{\bm{M}}}

\def\mX{{\bm{X}}}

\DeclareMathAlphabet{\mathsfit}{\encodingdefault}{\sfdefault}{m}{sl}
\SetMathAlphabet{\mathsfit}{bold}{\encodingdefault}{\sfdefault}{bx}{n}

\def\gE{{\mathcal{E}}}

\def\gG{{\mathcal{G}}}

\def\gV{{\mathcal{V}}}

\def\sA{{\mathbb{A}}}

\def\sR{{\mathbb{R}}}

\def\sZ{{\mathbb{Z}}}

\definecolor{mydarkblue}{rgb}{0,0.08,0.45}

\usepackage[pagebackref=true,breaklinks=true,colorlinks,bookmarks=false,citecolor=mydarkblue, linkcolor=mydarkblue]{hyperref}
\usepackage{url}
\usepackage[version=4]{mhchem}
\usepackage{graphicx}
\usepackage{wrapfig}
\usepackage{siunitx}
\usepackage{algorithm}
\usepackage[noend]{algpseudocode}
\usepackage{multirow}
\usepackage{tablefootnote}
\usepackage{amsthm}
\usepackage{placeins}

\newcommand*\diff{\mathop{}\!\mathrm{d}}

\usepackage[compact]{titlesec}
\def\setstretch#1{\renewcommand{\baselinestretch}{#1}}
\usepackage{etoolbox}
\makeatletter
\patchcmd{\hyper@makecurrent}{%
    \ifx\Hy@param\Hy@chapterstring
        \let\Hy@param\Hy@chapapp
    \fi
}{%
    \iftoggle{inappendix}{
        \@checkappendixparam{chapter}%
        \@checkappendixparam{section}%
        \@checkappendixparam{subsection}%
        \@checkappendixparam{subsubsection}%
        \@checkappendixparam{paragraph}%
        \@checkappendixparam{subparagraph}%
    }{}%
}{}{\errmessage{failed to patch}}

\newcommand*{\@checkappendixparam}[1]{%
    \def\@checkappendixparamtmp{#1}%
    \ifx\Hy@param\@checkappendixparamtmp
        \let\Hy@param\Hy@appendixstring
    \fi
}
\makeatletter

\newtoggle{inappendix}
\togglefalse{inappendix}

\apptocmd{\appendix}{\toggletrue{inappendix}}{}{\errmessage{failed to patch}}

\providecommand{\REVISION}[1]{{#1 }}
\title{Crystal Diffusion Variational Autoencoder for Periodic Material Generation}

\author{Tian Xie\thanks{Equal contribution. Correspondence to: Tian Xie at txie@csail.mit.edu}, Xiang Fu\footnotemark[1], Octavian-Eugen Ganea\footnotemark[1], Regina Barzilay,  Tommi Jaakkola  \\
Computer Science and Artificial Intelligence Laboratory\\
Massachusetts Institute of Technology\\
Cambridge, MA 02139, USA \\
\texttt{\{txie,xiangfu,oct,regina,tommi\}@csail.mit.edu} \\
}

\begin{document}

\maketitle

\begin{abstract}
Generating the periodic structure of stable materials is a long-standing challenge for the material design community. This task is difficult because stable materials only exist in a low-dimensional subspace of all possible periodic arrangements of atoms: 1) the coordinates must lie in the local energy minimum defined by quantum mechanics, and 2) global stability also requires the structure to follow the complex, yet specific bonding preferences between different atom types. Existing methods fail to incorporate these factors and often lack proper invariances. We propose a Crystal Diffusion Variational Autoencoder (CDVAE) that captures the physical inductive bias of material stability. By learning from the data distribution of stable materials, the decoder generates materials in a diffusion process that moves atomic coordinates towards a lower energy state and updates atom types to satisfy bonding preferences between neighbors. Our model also explicitly encodes interactions across periodic boundaries and respects permutation, translation, rotation, and periodic invariances. We significantly outperform past methods in three tasks: 1) reconstructing the input structure, 2) generating valid, diverse, and realistic materials, and 3) generating materials that optimize a specific property. We also provide several standard datasets and evaluation metrics for the broader machine learning community. \footnote{Code and data are available at \url{https://github.com/txie-93/cdvae}}
\end{abstract}

\section{Introduction}

\begin{wrapfigure}{tr}{0.6\textwidth}
\vspace*{-3.3ex}
  \begin{center}
    \includegraphics[width=0.6\textwidth]{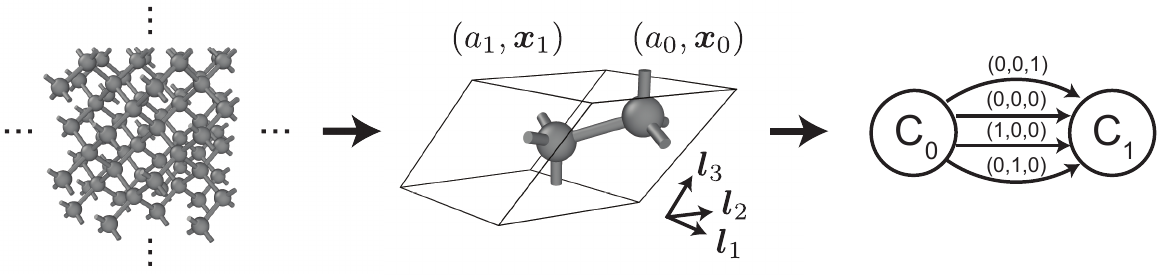}
  \end{center}
  \vspace*{-3.3ex}
  \caption{The periodic structure of diamond. The left shows the infinite periodic structure, the middle shows a unit cell representing the periodic structure, and the right shows a multi-graph \citep{xie2018crystal} representation.}
  \vspace*{-4ex}
  \label{fig:diamond}
\end{wrapfigure}

Solid state materials, represented by the periodic arrangement of atoms in the 3D space, are the foundation of many key technologies including solar cells, batteries, and catalysis \citep{butler2018machine}. Despite the rapid progress of molecular generative models and their significant impact on drug discovery, the problem of material generation has many unique challenges. Compared with small molecules, materials have more complex periodic 3D structures and cannot be adequately represented by a simple graph like molecular graphs (\autoref{fig:diamond}). In addition, materials can be made up of more than 100 elements in the periodic table, while molecules are generally only made up of a small subset of atoms such as carbon, oxygen, and hydrogen. Finally, the data for training ML models for material design is limited. There are only $\sim$200k experimentally known inorganic materials, collected by the ICSD \citep{belsky2002new}, in contrast to close to a billion molecules in ZINC \citep{irwin2005zinc}.

The key challenge of this task is in generating \textit{stable} materials. Such materials only exist in a low-dimensional subspace of all possible periodic arrangements of atoms: 1) the atom coordinates must lie in the local energy minimum defined by quantum mechanics (QM); 2) global stability also requires the structure to follow the complex, yet specific bonding preferences between different atom types (\autoref{sec:problem}). The issue of stability is unique to material generation because valency checkers assessing molecular stability are not applicable to materials. Moreover, we also have to encode the interactions crossing periodic boundaries (\autoref{fig:diamond}, middle), and satisfy permutation, translation, rotation, and periodic invariances (\autoref{sec:material_representation}). Our goal is to learn representations that can learn features of stable materials from data, while adhering to  
the above invariance properties.

We address these challenges by learning a variational autoencoder (VAE) \citep{corr/KingmaW13} to generate stable 3D materials directly from a latent representation without intermediates like graphs. The key insight is to exploit the fact that all materials in the data distribution are stable, therefore if noise is added to the ground truth structure, denoising it back to its original structure will likely increase stability. We capture this insight by designing a noise conditional score network (NCSN) \citep{nips/SongE19} as our decoder: 1) the decoder outputs gradients that drive the atom coordinates to the energy local minimum; 2) it also updates atom types based on the neighbors to capture the specific local bonding preferences (e.g., Si-O is preferred over Si-Si and O-O in \ce{SiO2}). During generation, materials are generated using Langevin dynamics that gradually deforms an initial random structure to a stable structure. To capture the necessary invariances and encode the interactions crossing periodic boundaries, we use SE(3) equivariant graph neural networks adapted with periodicity (PGNNs) for both the encoder and decoder of our VAE.

Our theoretical analysis further reveals an intriguing connection between the gradient field learned by our decoder and an harmonic force field. De facto, the decoder utilizes the latter to estimate the forces on atoms when their coordinates deviate from the equilibrium positions. Consequently, this formulation provides an important physical inductive bias for generating stable materials.

In this work, we propose Crystal Diffusion Variational AutoEncoder (CDVAE) to generate stable materials by learning from the data distribution of known materials. Our main contributions include:
\begin{itemize}
    \item We curate 3 standard datasets from QM simulations and create a set of physically meaningful tasks and metrics for the problem of material generation.
    \item We incorporate stability as an inductive bias by designing a noise conditional score network as the decoder of our VAE, which allows us to generate significantly more realistic materials.
    \item We encode permutation, translation, rotation, and periodic invariances, as well as interactions crossing periodic boundaries with SE(3) equivariant GNNs adapted with periodicity.
    \item Empirically, our model significantly outperforms past methods in tasks including reconstructing an input structure, generating valid, diverse, and realistic materials, and generating materials that optimize specific properties.
\end{itemize}

\section{Related Work} \label{sec:related-work}

\textbf{Material graph representation learning.} Graph neural networks have made major impacts in material property prediction. They were first applied to the representation learning of periodic materials by \cite{xie2018crystal} and later enhanced by many studies including \cite{schutt2018schnet,chen2019graph}. The Open Catalyst Project (OCP) provides a platform for comparing different architectures by predicting energies and forces from the periodic structure of catalytic surfaces \citep{chanussot2021open}. Our encoder and decoder PGNNs directly use GNN architectures developed for the OCP \citep{iclr/KlicperaGG20,klicpera2021gemnet,shuaibi2021rotation,godwin2021very}, which are also closely related to SE(3) equivariant networks \citep{thomas2018tensor,nips/FuchsW0W20}.

\textbf{Quantum mechanical search of stable materials.} Predicting the structure of unknown materials requires very expensive random search and QM simulations, and is considered a grand challenge in materials discovery \citep{oganov2019structure}. State-of-the-art methods include random sampling \citep{pickard2011ab}, evolutionary algorithms \citep{wang2012calypso, glass2006uspex}, substituting elements in known materials \citep{hautier2011data}, etc., but they generally have low success rates and require extensive computation even on relatively small problems.

\textbf{Material generative models.} Past material generative models mainly focus on two different approaches, and neither incorporate stability as an inductive bias. The first approach treats materials as 3D voxel images, but the process of decoding images back to atom types and coordinates often results in low validity, and the models are not rotationally invariant \citep{hoffmann2019data,noh2019inverse,court20203,long2021constrained}. The second directly encodes atom coordinates, types, and lattices as vectors \citep{ren2020inverse,kim2020generative,zhao2021high}, but the models are generally not invariant to any Euclidean transformations. Another related method is to train a force field from QM forces and then apply the learned force field to generate stable materials by minimizing energy \citep{deringer2018data, chen2022universal}. This method is conceptually similar to our decoder, but it requires additional force data which is expensive to obtain. Remotely related works include generating contact maps from chemical compositions \citep{hu2021contact,yang2021crystal} and building generative models only for chemical compositions \citep{sawada2019study,pathak2020deep,dan2020generative}. 

\textbf{Molecular conformer generation \REVISION{and protein folding}.} Our decoder that generates the 3D atomic structures via a diffusion process is closely related to the diffusion models used for molecular conformer generation \citep{icml/ShiLXT21,xu2021geodiff}. The key difference is that our model does not rely on intermediate representations like molecular graphs. G-SchNet \citep{gebauer2019symmetry} is more closely related to our method because it directly generates 3D molecules atom-by-atom without relying on a graph. Another closely related work is E-NFs \citep{satorras2021n} that use a flow model to generate 3D molecules. \REVISION{In addition, score-based and energy-based models have also been used for molecular graph generation \citep{liu2021graphebm} and protein folding \citep{wu2021se}. Flow models have also been used for molecular graph generation \citep{shi2020graphaf,luo2021graphdf}.  However, these generative models do not incorporate periodicity}, which makes them unsuitable for materials.

\section{Preliminaries}

\subsection{Periodic structure of materials} \label{sec:material_representation}

Any material structure can be represented as the periodic arrangement of atoms in the 3D space. As illustrated in \autoref{fig:diamond}, we can always find a repeating unit, i.e. a unit cell, to describe the infinite periodic structure of a material. A unit cell that includes $N$ atoms can be fully described by 3 lists: 1) atom types $\mA = (a_0, ..., a_N) \in \sA^{N}$, where $\sA$ denotes the set of all chemical elements; 2) atom coordinates $\mX = (\vx_0, ..., \vx_N) \in \sR^{N \times 3}$; and 3) periodic lattice $\mL = (\vl_1, \vl_2, \vl_3) \in \sR^{3 \times 3}$. The periodic lattice defines the periodic translation symmetry of the material. Given $\mM = (\mA,\mX,\mL)$, the infinite periodic structure can be represented as,
\begin{equation}
    \{ (a_i', \vx_i') | a_i' = a_i, \vx_i' = \vx_i + k_1 \vl_1 + k_2 \vl_2 + k_3 \vl_3, k_1, k_2, k_3 \in \sZ \},
\end{equation}
where $k_1,k_2,k_3$ are any integers that translate the unit cell using $\mL$ to tile the entire 3D space.

The chemical composition of a material denotes the ratio of different elements that the material is composed of. Given the atom types of a material with $N$ atoms $\mA \in \sA^N$, the composition can be represented as $\vc \in \sR^{|\sA|}$, where $\vc_i > 0$ denotes the percentage of atom type $i$ and $\sum_i \vc_i = 1$. For example, the composition of diamond in \autoref{fig:diamond} has $\vc_6 = 1$ and $\vc_i = 0$ for $i\neq 6$ because 6 is the atomic number of carbon.

\textbf{Invariances for materials.} The structure of a material does not change under several invariances. 1) \textit{Permutation invariance}. Exchanging the indices of any pair of atoms will not change the material. 2) \textit{Translation invariance}. Translating the atom coordinates $\mX$ by an arbitrary vector will not change the material. 3) \textit{Rotation invariance}. Rotating $\mX$ and $\mL$ together by an arbitrary rotation matrix will not change the material. 4) \textit{Periodic invariance}. There are infinite different ways of choosing unit cells with different shapes and sizes, e.g., obtaining a bigger unit cell as an integer multiplier of a smaller unit cell using integer translations. The material will again not change given different choices of unit cells.

\textbf{Multi-graph representation for materials.} Materials can be represented as a directed multi-graph $\gG=\{\gV, \gE\}$ to encode the periodic structures following \citep{wells1977three,10.2307/36648,xie2018crystal}, where $\gV = \{v_1, ..., v_N\}$ is the set of nodes representing atoms and $\gE = \{e_{ij, (k_1, k_2, k_3)} | i, j \in \{1, ..., N\}, k_1, k_2, k_3 \in \sZ \}$ is the set of edges representing bonds. $e_{ij, (k_1, k_2, k_3)}$ denotes a directed edge from node $i$ at the original unit cell to node $j$ at the cell translated by $k_1 \vl_1 + k_2 \vl_2 + k_3 \vl_3$ (in \autoref{fig:diamond} right, $(k_1, k_2, k_3)$ are labeled on top of edges). For materials, there is no unique way to define edges (bonds) and the edges are often computed using k-nearest neighbor (KNN) approaches under periodicity or more advanced methods such as CrystalNN \citep{pan2021benchmarking}. Given this directed multi-graph, message-passing neural networks and SE(3)-equivariant networks can be used for the representation learning of materials.


\subsection{Problem definition and its physical origin} \label{sec:problem}

Our goal is to generate novel, stable materials $\mM = (\mA,\mX,\mL) \in \sA^{N} \times \sR^{N \times 3} \times \sR^{3 \times 3}$. The space of stable materials is a subspace in $\sA^{N} \times \sR^{N \times 3} \times \sR^{3 \times 3}$ that satisfies the following constraints. 1) The materials lie in the local minimum of the energy landscape defined by quantum mechanics, with respect to the atom coordinates and lattice, i.e. $\partial E / \partial \mX = \bm{0}$ and $\partial E / \partial \mL = \bm{0}$. 2) The material is globally stable and thus cannot decompose into nearby phases. Global stability is strongly related to bonding preferences between neighboring atoms. For example, in \ce{SiO2}, each Si is surrounded by 4 O and each O is surrounded by 2 Si. This configuration is caused by the stronger bonding preferences between Si-O than Si-Si and O-O. 

Generally, finding novel, stable materials requires very expensive random search and quantum mechanical simulations. To bypass this challenge, we aim to learn a generative model $p(\mM)$ from the empirical distribution of experimentally observed stable materials. A successful generative model will be able to generate novel materials that satisfy the above constraints, which can then be verified using quantum mechanical simulations.

\subsection{Diffusion models}

Diffusion models are a new class of generative models that have recently shown great success in generating high-quality images \citep{dhariwal2021diffusion}, point clouds \citep{cai2020learning,luo2021diffusion}, and molecular conformations \citep{icml/ShiLXT21}. There are several different types of diffusion models including diffusion probabilistic models \citep{sohl2015deep}, noise-conditioned score networks (NCSN) \citep{nips/SongE19}, and denoising diffusion probabilistic models (DDPM) \citep{nips/HoJA20}. We follow ideas from the NCSN \citep{nips/SongE19} and learn a score network $\vs_{\bm{\theta}}(\vx)$ to approximate the gradient of a probability density $\nabla_\vx p(\vx)$ at different noise levels. Let $\{ \sigma_i \}_{i=1}^L$ be a  sequence of positive scalars that satisfies $\sigma_1 / \sigma_2 = ... = \sigma_{L-1} / \sigma_L > 1$. We define the data distribution perturbed by Gaussian noise $\sigma$ as $q_\sigma(\vx) = \int p_{\mathrm{data}}(\vt) \mathcal{N} (\vx | \vt, \sigma^2 I) \diff \vt$. The goal of NCSN is to learn a score network to jointly estimate the scores of all perturbed data distributions, i.e. $\forall \sigma \in \{ \sigma_i \}_{i=1}^L: \vs_{\bm{\theta}}(\vx, \sigma) \approx \nabla_\vx q_\sigma(\vx)$. During generation, NCSN uses an annealed Langevin dynamics algorithm to produce samples following the gradient estimated by the score network with a gradually reduced noise level.

\section{Proposed Method} \label{sec:method}

\begin{figure}
    \centering
    \includegraphics[width=0.9\textwidth]{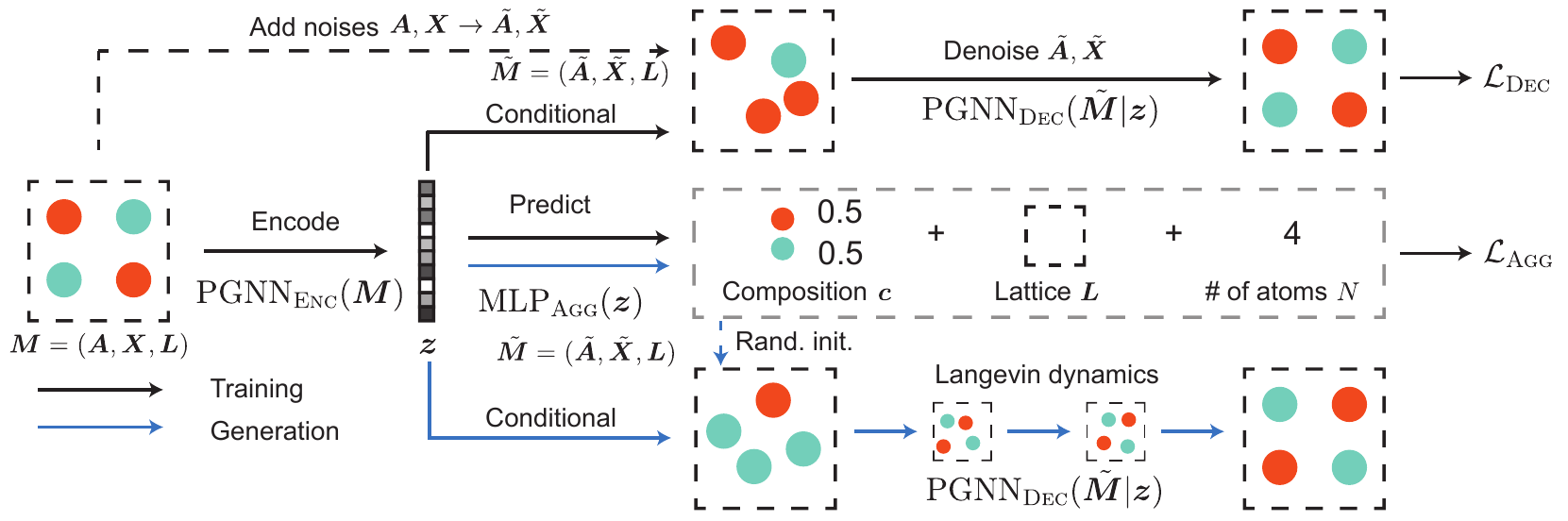}
    \caption{Overview of the proposed CDVAE approach.}
    \label{fig:illustrative}
\end{figure}

Our approach generates new materials via a two-step process: 1) We sample a $\vz$ from the latent space and use it to predict 3 aggregated properties of a material: composition ($\vc$), lattice ($\mL$), and number of atoms ($N$), which are then used to randomly initialize a material structure $\tilde{\mM} = (\tilde{\mA}, \tilde{\mX}, \mL)$. 2) We perform Langevin dynamics to simultaneously denoise $\tilde{\mX}$ and $\tilde{\mA}$ conditioned on $\vz$ to improve both the local and global stability of $\tilde{\mM}$ and generate the final structure of the new material.

To train our model, we optimize 3 networks concurrently using stable materials $\mM = (\mA, \mX, \mL)$ sampled from the data distribution. 1) A periodic GNN encoder $\mathrm{PGNN}_{\textsc{Enc}}(\mM)$ that encodes $\mM$ into a latent representation $\vz$. 2) A property predictor $\mathrm{MLP}_{\textsc{Agg}}(\vz)$ that predicts the $\vc$, $\mL$, and $N$ of $\mM$ from $\vz$. 3) A periodic GNN decoder $\mathrm{PGNN}_{\textsc{Dec}}(\tilde{\mM} | \vz)$ that denoises both $\tilde{\mX}$ and $\tilde{\mA}$ conditioned on $\vz$. For 3), the noisy structure $\tilde{\mM} = (\tilde{\mA}, \tilde{\mX}, \mL)$ is obtained by adding different levels of noise to $\mX$ and $\mA$. The noise schedules are defined by the predicted aggregated properties, with the motivation of simplifying the task for our decoder from denoising an arbitrary random structure from over $\sim$100 elements to a constrained random structure from predicted properties. We train all three networks together by minimizing a combined loss including the aggregated property loss $\mathcal{L}_{\textsc{Agg}}$, decoder denoising loss $\mathcal{L}_{\textsc{Dec}}$, and a KL divergence loss $\mathcal{L}_{\textsc{KL}}$ for the VAE.

To capture the interactions across periodic boundaries, we employ a multi-graph representation (\autoref{sec:material_representation}) for both $\mM$ and $\tilde{\mM}$. We also use SE(3) equivariant GNNs adapted with periodicity as both the encoder and the decoder to ensure the permutation, translation, rotation, and periodic invariances of our model. The CDVAE is summarized in \autoref{fig:illustrative} and we explain the individual components of our method below. The implementation details can be found in \autoref{sec:implementation}.

\textbf{Periodic material encoder.}
$\mathrm{PGNN}_{\textsc{Enc}}(\mM)$ encodes a material $\mM$ as a latent representation $\vz \in \sR^D$ following the reparameterization trick in VAE \citep{corr/KingmaW13}. We use the multi-graph representation (refer to \autoref{sec:material_representation}) to encode $\mM$, and $\mathrm{PGNN}_{\textsc{Enc}}$ can be parameterized with an SE(3) invariant graph neural network.

\textbf{Prediction of aggregated properties.} $\mathrm{MLP}_{\textsc{Agg}}(\vz)$ predicts 3 aggregated properties of the encoded material from its latent representation $\vz$. It is parameterized by 3 separate multilayer perceptrons (MLPs). 1) Composition $\vc \in \sR^{|\sA|}$ is predicted by minimizing the cross entropy between the ground truth composition and predicted composition, i.e. $-\sum_i \vp_i \log \vc_i$. 2) Lattice $\mL \in \sR^{3 \times 3}$ is reduced to 6 unique, rotation invariant parameters with the Niggli algorithm \citep{grosse2004numerically}, i.e., the lengths of the 3 lattice vectors, the angles between them, and the values are predicted with an MLP after being normalized to the same scale (\autoref{sec:lattice-prediction}) with an $L_2$ loss. 3) Number of atoms $N \in \{1, 2, ...\}$ is predicted with a softmax classification loss from the set of possible number of atoms. $\mathcal{L}_{\textsc{Agg}}$ is a weighted sum of the above 3 losses.

\textbf{Conditional score matching decoder.} $\mathrm{PGNN}_{\textsc{Dec}}(\tilde{\mM} | \vz)$ is a PGNN that inputs a noisy material $\tilde{\mM}$ with type noises $\sigma_{\mA}$, coordinate noises $\sigma_{\mX}$, as well as a latent $\vz$, and outputs 1) a score $\vs_{\mX}(\tilde{\mM} | \vz; \sigma_{\mA}, \sigma_{\mX}) \in \sR^{N \times 3}$ to denoise the coordinate for each atom towards its ground truth value, and 2) a probability distribution of the true atom types $\vp_{\mA}(\tilde{\mM} | \vz; \sigma_{\mA}, \sigma_{\mX}) \in \sR^{N \times |\sA|}$. We use a SE(3) graph network to ensure the equivariance of $\vs_\mX$
with respect to the rotation of $\tilde{\mM}$. To obtain the noisy structures $\tilde{\mM}$, we sample $\sigma_{\mA}$ and $\sigma_{\mX}$ from two geometric sequences of the same length: $\{ \sigma_{\mA,j} \}_{j=1}^L, \{ \sigma_{\mX,j} \}_{j=1}^L$, and add the noises with the following methods. For type noises, we use the type distribution defined by the predicted composition $\vc$ to linearly perturb true type distribution \REVISION{$\tilde{\mA} \sim (\frac{1}{1 + \sigma_{\mA}} \vp_\mA + \frac{\sigma_{\mA}}{1 + \sigma_{\mA}} \vp_\vc)$}, where $\vp_{\mA,ij} = 1$ if atom $i$ has the true atom type $j$ and $\vp_{\mA,ij} = 0$ for all other $j$s, and $\vp_\vc$ is the predicted composition. For coordinate noises, we add Gaussian noises to the true coordinates $\tilde{\mX} \sim \mathcal{N}(\mX, \sigma_{\mX}^2 \mI)$.

$\mathrm{PGNN}_{\textsc{Dec}}$ is parameterized by a SE(3) equivariant PGNN that inputs a multi-graph representation (\autoref{sec:material_representation}) of the noisy material structure and the latent representation. The node embedding for node $i$ is obtained by the concatenation of the element embedding of $\tilde{a}_i$ and the latent representation $\vz$, followed by a MLP, $\vh_i^0 = \mathrm{MLP}(\ve_{\mathrm{a}}(\tilde{a}_i) \mathbin\Vert \vz)$,
where $\mathbin\Vert$ denotes concatenation of two vectors and $\ve_{\mathrm{a}}$ is a learned embedding for elements. After $K$ message-passing layers, $\mathrm{PGNN}_{\textsc{Dec}}$ outputs a vector per node that is equivariant to the rotation of $\tilde{\mM}$. These vectors are used to predict the scores, and we follow \cite{nips/SongE19,icml/ShiLXT21} to parameterize the score network with noise scaling: $\vs_\mX(\tilde{\mM} | \vz; \sigma_{\mA}, \sigma_{\mX}) = \vs_\mX(\tilde{\mM} | \vz) / \sigma_{\mX}$. The node representations $\vh_i^K$ are used to predict the distribution of true atom types, and the type predictor is the same at all noise levels: $\vp_\mA(\tilde{\mM} | \vz; \sigma_{\mA}, \sigma_{\mX}) = \vp_\mA(\tilde{\mM} | \vz)$, $\vp_\mA(\tilde{\mM} | \vz)_{i} = \mathrm{softmax}(\mathrm{MLP}(\vh_i^K))$.

\textbf{Periodicity influences denoising target.} Due to periodicity, a specific atom $i$ may move out of the unit cell defined by $\mL$ when the noise is sufficiently large. This leads to two different ways to define the scores for node $i$. 1) Ignore periodicity and define the target score as $\vx_i - \tilde{\vx}_i$; or 2) Define the target score as the shortest possible displacement between $\vx_i$ and $\tilde{\vx}_i$ considering periodicity, i.e. $\vd_{\mathrm{min}}(\vx_i,\tilde{\vx}_i) = \min_{k_1, k_2, k_3} (\vx_i - \tilde{\vx}_i + k_1 \vl_1 + k_2 \vl_2 + k_3 \vl_3)$. We choose 2) because the scores are the same given two different $\tilde{\mX}$ that are periodically equivalent, which is mathematically grounded for periodic structures, and empirically results in much more stable training.

The training loss for the decoder $\mathcal{L}_{\textsc{Dec}}$ can be written as,
\begin{equation}\label{eq:loss}\small
    \frac{1}{2L} \sum_{j=1}^L \left[ \mathbb{E}_{q_{\mathrm{data}(\mM)}} \mathbb{E}_{q_{\sigma_{\mA,j},\sigma_{\mX,j}}(\tilde{\mM} | \mM )} \left( \Big\lVert \vs_\mX (\tilde{\mM} | \vz) - \frac{\vd_{\mathrm{min}}(\mX, \tilde{\mX})}{\sigma_{\mX,j}} \Big\rVert_2^2 + \frac{\lambda_{\mathrm{a}}}{\sigma_{\mA,j}} \mathcal{L}_{\mathrm{a}}(\vp_\mA(\tilde{\mM} | \vz), \vp_\mA) \right)  \right],
\end{equation}
where $\lambda_{\mathrm{a}}$ denotes a coefficient for balancing the coordinate and type losses, $\mathcal{L}_{\mathrm{a}}$ denotes the cross entropy loss over atom types, $\vp_\mA$ denotes the true atom type distribution. Note that to simplify the equation, we follow the loss coefficients in \cite{nips/SongE19} for different $\sigma_{\mX,j}$ and $\sigma_{\mA,j}$ and factor them into \autoref{eq:loss}.

\begin{wrapfigure}{r}{0.55\textwidth}
    \vspace*{-5.5ex}
    \begin{minipage}{0.55\textwidth}
    \begin{algorithm}[H]
    \small
    \caption{Material Generation via Annealed Langevin Dynamics}\label{algorithm}
    \begin{algorithmic}[1]
    \State \textbf{Input:} latent representation $\vz$, type and coordinate noise levels $\{ \sigma_\mA \}$, $\{ \sigma_\mX \}$, step size $\epsilon$, number of sampling steps $T$
    \State Predict aggregated properties $\vc, \mL, N$ from $\vz$.
    \State Uniformly initialize $\mX_0$ within the unit cell by $\mL$. \State Randomly initialize $\mA_0$ with $\vc$. 
    
    \For{$j \gets 1$ to $L$}
        \State{$\alpha_j \gets \epsilon \cdot \sigma_{\mX,j}^2/\sigma_{\mX,L}^2$}
        \For{$t \gets 1$ to $T$}
            \State{$\vs_{\mX,t} \gets \vs_\mX(\mA_{t-1}, \mX_{t-1}, \mL | \vz; \sigma_{\mA,j}, \sigma_{\mX,j})$}
            \State{$\vp_{\mA,t} \gets \vp_\mA(\mA_{t-1}, \mX_{t-1}, \mL | \vz; \sigma_{\mA,j}, \sigma_{\mX,j})$}
            \State{Draw $\mX_t^\epsilon \sim \mathcal{N}(0, \mI)$}
            \State{$\mX_t' \gets \mX_{t-1} + \alpha_j \vs_{\mX,t} + \sqrt{2\alpha_i}\mX_t^\epsilon$}
            \State{$\mX_t \gets \operatorname{back\_to\_cell}(\mX_t', \mL)$}
            \State{$\mA_t = \operatorname{argmax}\vp_{\mA,t}$}
        \EndFor
        \State{$\mX_0 \gets \mX_T, \mA_0 \gets \mA_T$}
    \EndFor
    
    \end{algorithmic}
    \end{algorithm}
    \end{minipage}
    \vspace*{-2ex}
\end{wrapfigure}

\textbf{Material generation with Langevin dynamics.} After training the model, we can generate the periodic structure of material given a latent representation $\vz$. First, we use $\vz$ to predict the aggregated properties: 1) composition $\vc$, 2) lattice $\mL$, and 3) the number of atoms $N$. Then, we randomly initialize an initial periodic structure $(\mA_0, \mX_0, \mL)$ with the aggregated properties and perform an annealed Langevin dynamics \citep{nips/SongE19} using the decoder, simultaneously updating the atom types and coordinates. During the coordinate update, we map the coordinates back to the unit cell at each step if atoms move out of the cell. The algorithm is summarized in Algorithm \ref{algorithm}.

\textbf{Connection between the gradient field and a harmonic force field.} The gradient field $\vs_\mX(\tilde{\mM} | \vz)$ is used to update atom coordinates in Langevin dynamics via the force term, $\alpha_j \vs_{\mX, t}$. In \autoref{sec:proof}, we show that $\alpha_j \vs_{\mX, t}$ is mathematically equivalent to\footnote{In fact, this is also true for the original formulation of NCSN \citep{nips/SongE19}} a harmonic force field $F(\tilde{\mX}) = -k (\tilde{\mX} - \mX)$ when the noises are small, where $\mX$ is the equilibrium position of the atoms and $k$ is a force constant. Harmonic force field, i.e. spring-like force field, is a simple yet general physical model that approximates the forces on atoms when they are close to their equilibrium locations. This indicates that our learned gradient field utilizes the harmonic approximation to approximate QM forces without any explicit force data and generates stable materials with this physically motivated inductive bias.



\section{Experiments}

We evaluate multiple aspects of material generation that are related to real-world material discovery process. Past studies in this field used very different tasks and metrics, making it difficult to compare different methods. Building upon past studies \citep{court20203,ren2020inverse}, we create a set of standard tasks, datasets, and metrics to evaluate and compare models for material generation. Experiment details can be found in \autoref{sec:experimental-details}.

\textbf{Tasks.} We focus on 3 tasks for material generation. 1) \textit{Reconstruction} evaluates the ability of the model to reconstruct the original material from its latent representation $\vz$. 2) \textit{Generation} evaluates the validity, property statistics, and diversity of material structures generated by the model. 3) \textit{Property optimization} evaluates the model's ability to generate materials that are optimized for a specific property.

\textbf{Datasets.} We curated 3 datasets representing different types of material distributions. 1) \textbf{Perov-5} \citep{castelli2012new,castelli2012computational} includes 18928 perovskite materials that share the same structure but differ in composition. There are 56 elements and all materials have 5 atoms in the unit cell. 2) \textbf{Carbon-24} \citep{carbon2020data} includes 10153 materials that are all made up of carbon atoms but differ in structures. There is 1 element and the materials have 6 - 24 atoms in the unit cells. 3) \textbf{MP-20} \citep{jain2013commentary} includes 45231 materials that differ in both structure and composition. There are 89 elements and the materials have 1 - 20 atoms in the unit cells. We use a 60-20-20 random split for all of our experiments. Details regarding dataset curation can be found at \autoref{app-sec:dataset}.

\textbf{Stability of materials in datasets.} Structures in all 3 datasets are obtained from QM simulations and all structures are at local energy  minima. Most materials in Perov-5 and Carbon-24 are hypothetical, i.e. they may not have global stability (\autoref{sec:problem}) and likely cannot be synthesized. MP-20 is a realistic dataset that includes most experimentally known inorganic materials with at most 20 atoms in the unit cell, most of which are globally stable. A model achieving good performance in MP-20 has the potential to generate novel materials that can be experimentally synthesized.

\textbf{Baselines.} We compare CDVAE with the following \REVISION{4} baselines, which include the latest coordinate-based, voxel-based, and 3D molecule generation methods.
\textbf{FTCP}~\citep{ren2020inverse} is a crystal representation that concatenates real-space properties (atom positions, atom types, etc.) and Fourier-transformed momentum-space properties (diffraction pattern). A 1D CNN-VAE is trained over this representation for crystal generation.
%
%
\textbf{Cond-DFC-VAE}~\citep{court20203} encodes and generates crystals with 3D density maps, while employing several modifications over the previous Voxel-VAE \citep{hoffmann2019data} method. However, the effectiveness is only demonstrated for cubic systems, limiting its usage to the Perov-5 dataset.
\textbf{G-SchNet}~\citep{gebauer2019symmetry} is an auto-regressive model that generates 3D molecules by performing atom-by-atom completion using SchNet \citep{schutt2018schnet}. Since G-SchNet is unaware of periodicity and cannot generate the lattice $\mL$. We adapt G-SchNet to our material generation tasks by constructing the smallest oriented bounding box with PCA such that the introduced periodicity does not cause structural invalidity.
\REVISION{\textbf{P-G-SchNet} is our modified G-SchNet that incorporates periodicity. During training, the SchNet encoder inputs the partial periodic structure to predict next atoms. During generation, we first randomly sample a lattice $\mL$ from training data and autoregressively generate the periodic structure.}

\subsection{Material reconstruction}

\begin{figure}
    \centering
    \includegraphics[width=0.7\textwidth]{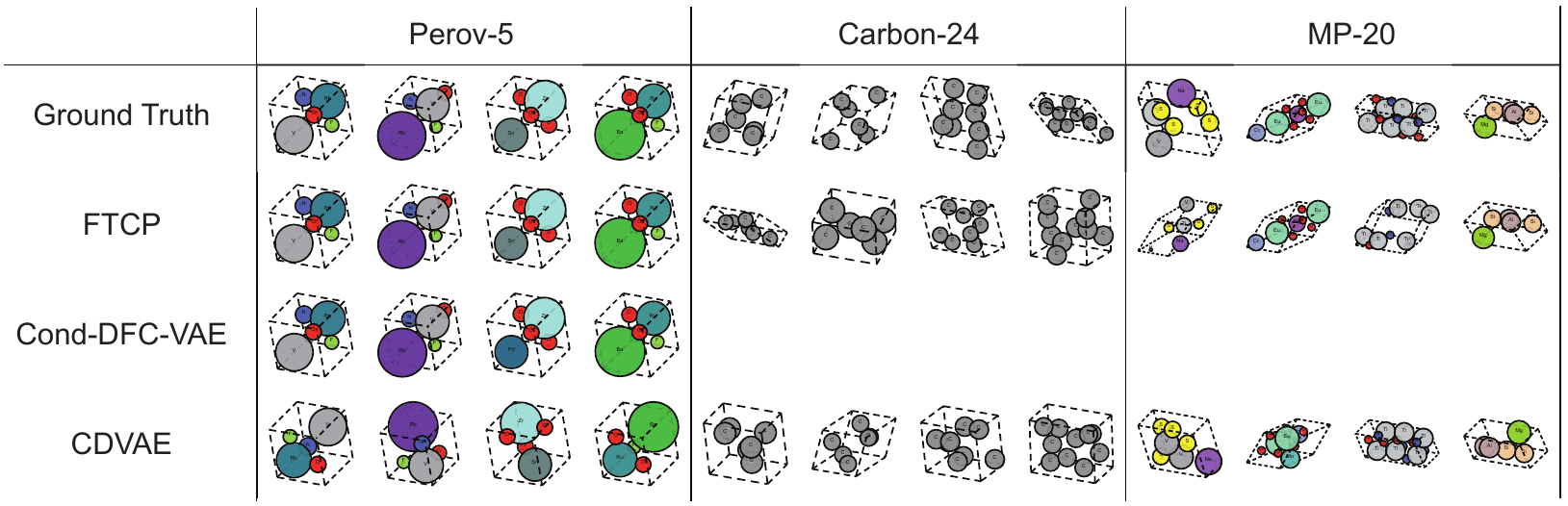}
    \caption{Reconstructed structures of randomly selected materials in the test set. Note our model reconstructs rotated (translated) version of the original material due to the SE(3) invariance.}
    \label{fig:reconstruction}
\end{figure}

\begin{table}[t]
\caption{Reconstruction performance.}
\scriptsize
\label{table:reconstruction}
\begin{center}
\begin{tabular}{l|llllll}
\multirow{2}{*}{\bf Method}  &\multicolumn{3}{c}{\bf Match rate (\%) $\uparrow$} & \multicolumn{3}{c}{\bf RMSE $\downarrow$} \\
 & \multicolumn{1}{c}{Perov-5} & \multicolumn{1}{c}{Carbon-24} & \multicolumn{1}{c}{MP-20} & \multicolumn{1}{c}{Perov-5} & \multicolumn{1}{c}{Carbon-24} & \multicolumn{1}{c}{MP-20}\\
 \hline 
FTCP & \textbf{99.34} & \textbf{62.28} & \textbf{69.89} & 0.0259 & 0.2563 & 0.1593 \\
Cond-DFC-VAE & 51.65 & -- & -- & 0.0217 & -- & -- \\
CDVAE & 97.52 & 55.22 & 45.43 & \textbf{0.0156} & \textbf{0.1251} & \textbf{0.0356} \\
\end{tabular}
\end{center}
\end{table}

\textbf{Setup.} The first task is to reconstruct the material from its latent representation. We evaluate reconstruction performance by matching the generated structure and the input structure for all materials in the test set. We use \texttt{StructureMatcher} from \texttt{pymatgen} \citep{ong2013python}, which finds the best match between two structures considering all invariances of materials. The match rate is the percentage of materials satisfying the criteria \texttt{stol=0.5, angle\_tol=10, ltol=0.3}. The RMSE is averaged over all matched materials. Because the inter-atomic distances can vary significantly for different materials, the RMSE is normalized by $\sqrt[3]{V/N}$, roughly the average atom radius per material. Note G-SchNet is not a VAE so we do not evaluate its reconstruction performance.

\textbf{Results.} The reconstructed structures are shown in \autoref{fig:reconstruction} and the metrics are in \autoref{table:reconstruction}. Since our model is SE(3) invariant, the generated structures may be a translated (or rotated) version of the ground truth structure. Our model has a lower RMSE than all other models, indicating its stronger capability to reconstruct the original stable structures. FTCP has a higher match rate than our model. This can be explained by the fact that the same set of local structures can be assembled into different stable materials globally (e.g., two different crystal forms of ZnS). Our model is SE(3) invariant and only encodes local structures, while FTCP directly encodes the absolute coordinates and types of each atom. In \autoref{fig:sampling}, we show that CDVAE can generate different plausible arrangements of atoms by sampling 3 Langevin dynamics with different random seeds from the same $\vz$. We note that this capability could be an advantage since it generates more diverse structures than simply reconstructing the original ones.


\subsection{Material generation}

\begin{figure}
\REVISION{
    \centering
    \includegraphics[width=0.7\textwidth]{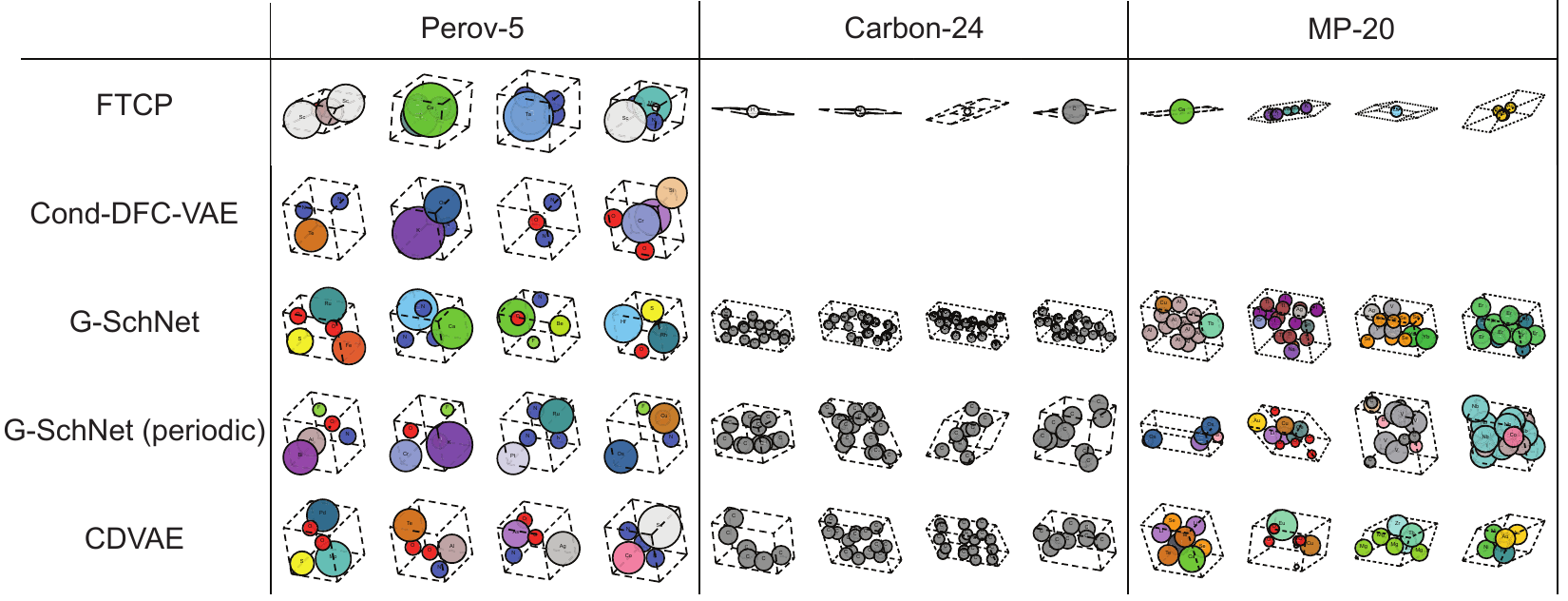}
    \caption{Structures sampled from $\mathcal{N}(0, 1)$ and filtered by the validity test.}
    \label{fig:generation}
}
    \vspace*{-2ex}
\end{figure}

\begin{table}[t]
\caption{Generation performance\tablefootnote{Some metrics unsuitable for specific datasets have ``--'' values in the table (explained in \autoref{sec:unsuitable-metrics}).}.}
\label{table:generation}
\scriptsize
\begin{center}
\begin{tabular}{ll|lllllllll}
\multirow{2}{*}{\bf Method} & \multirow{2}{*}{\bf Data} & \multicolumn{2}{c}{\bf Validity (\%) \tablefootnote{\REVISION{For comparison, the ground truth structure validity is 100.0 \% for all datasets, and ground truth composition validity is 98.60 \%, 100.0 \%, 91.13 \% for Perov-5, Carbon-24, and MP-20.}} $\uparrow$}  & \multicolumn{2}{c}{\bf COV (\%) $\uparrow$}  & \multicolumn{3}{c}{\bf Property Statistics $\downarrow$}  \\
& & Struc. & Comp. & R. & P. & $\rho$ & $E$ & \# elem. \\
\hline
\multirow[t]{3}{*}{FTCP \tablefootnote{Due to the low validity of FTCP, we instead randomly generate 100,000 materials from $\mathcal{N}(0, 1)$ and use 1,000 materials from those valid ones to compute diversity and property statistics metrics.}} & Perov-5 & 0.24 & 54.24 & \REVISION{0.00} & \REVISION{0.00} & 10.27 & 156.0 & 0.6297 \\
& Carbon-24 & 0.08 & -- & \REVISION{0.00} & \REVISION{0.00} & 5.206 & 19.05 & -- \\
& MP-20 & 1.55 & 48.37 & \REVISION{4.72} & \REVISION{0.09} & 23.71 & 160.9 & 0.7363 & \\
\multirow[t]{1}{*}{Cond-DFC-VAE} & Perov-5 & 73.60 & 82.95 & \REVISION{73.92} & \REVISION{10.13} & 2.268 & 4.111 & 0.8373 \\
\multirow[t]{3}{*}{G-SchNet} & Perov-5 & 99.92 & \textbf{98.79} & \REVISION{0.18} & \REVISION{0.23} & 1.625 &	4.746 &	\textbf{0.03684} \\
& Carbon-24  & 99.94	 & --	  & \REVISION{0.00} & \REVISION{0.00} & 0.9427 & 1.320 &	-- \\
& MP-20 & 99.65 &	75.96 & \REVISION{38.33} & \REVISION{\textbf{99.57}} &	3.034 &	42.09 &	0.6411	\\

\multirow[t]{3}{*}{P-G-SchNet} & Perov-5 & \REVISION{79.63} & \REVISION{\textbf{99.13}} & \REVISION{0.37} & \REVISION{0.25} & \REVISION{0.2755} & \REVISION{1.388} & \REVISION{0.4552} &  \\
& Carbon-24  & \REVISION{48.39} & \REVISION{--} & \REVISION{0.00} & \REVISION{0.00} & \REVISION{1.533} & \REVISION{134.7} & \REVISION{--}  \\
& MP-20 & \REVISION{77.51} & \REVISION{76.40} & \REVISION{41.93} & \REVISION{\textbf{99.74}} & \REVISION{4.04} & \REVISION{2.448} & \REVISION{\textbf{0.6234}} \\

\multirow[t]{3}{*}{CDVAE} & Perov-5 & \textbf{100.0} & \textbf{98.59} & \REVISION{\textbf{99.45}} & \REVISION{\textbf{98.46}} & \textbf{0.1258} & \textbf{0.0264} & 0.0628 \\
& Carbon-24 & \textbf{100.0} & -- & \REVISION{\textbf{99.80}} & \REVISION{\textbf{83.08}} & \textbf{0.1407} & \textbf{0.2850} & -- \\
& MP-20 & \textbf{100.0} & \textbf{86.70} & \REVISION{\textbf{99.15}} & \REVISION{\textbf{99.49}} & \textbf{0.6875} & \textbf{0.2778} & 1.432  \\
\end{tabular}
\end{center}
\end{table}

\textbf{Setup.} The second task is to generate novel, stable materials that are distributionally similar to the test materials. The only high-fidelity evaluation of stability of generated materials is to perform QM calculations, but it is computationally prohibitive to use QM for computing evaluation metrics. We developed several physically meaningful metrics to evaluate the validity, property statistics, and diversity of generated materials. 1) \textit{Validity}. Following \cite{court20203}, a structure is valid as long as the shortest distance between any pair of atoms is larger than \SI{0.5}{\angstrom}, which is a relative weak criterion. The composition is valid if the overall charge is neutral as computed by \texttt{SMACT} \citep{davies2019smact}. \REVISION{2) \textit{Coverage (COV)}. Inspired by \cite{xu2021learning,ganea2021geomol}, we define two coverage metrics, COV-R (Recall) and COV-P (Precision), to measure the similarity between ensembles of generated materials and ground truth materials in test set. Intuitively, COV-R measures the percentage of ground truth materials being correctly predicted, and COV-P measures the percentage of predicted materials having high quality (details in \autoref{sec:coverage}).} 3) \textit{Property statistics}. We compute the earth mover's distance (EMD) between the property distribution of generated materials and test materials. We use density ($\rho$, unit \SI{}{g/cm^3}), energy predicted by an independent GNN ($E$, unit \SI{}{eV/atom}), and number of unique elements (\# elem.) as our properties. Validity \REVISION{and coverage} are computed over 10,000 materials randomly sampled from $\mathcal{N}(0, 1)$. Property statistics is computed over 1,000 valid materials randomly sampled from those that pass the validity test.

\textbf{Results.} The generated structures are shown in \autoref{fig:generation} and the metrics are in \autoref{table:generation}. \REVISION{Our model achieves a higher validity than FTCP, Cond-DFC-VAE, and P-G-SchNet, while G-SchNet achieves a similar validity as ours. The lower structural validity in P-G-SchNet than G-SchNet is likely due to the difficulty of avoiding atom collisions during the autoregressive generation inside a finite periodic box. On the contrary, our G-SchNet baseline constructs the lattice box after the 3D positions of all atoms are generated, and the construction explicitly avoids introducing invalidity. Furthermore, our model also achieves higher COV-R and COV-P than all other models, except in MP-20 our COV-P is similar to G-SchNet and P-G-SchNet. These results indicate that our model generates both diverse (COV-R) and high quality (COV-P) materials. More detailed results on the choice of thresholds for COV-R and COV-P, as well as additional metrics can be found in \autoref{sec:coverage}. Finally, our model also significantly outperforms all other models in the property statistics of density and energy, further confirming the high quality of generated materials.} We observe that our method tends to generate more elements in a material than ground truth, which explains the lower performance in the statistics of \# of elems.\ than G-SchNet. We hypothesize this is due to the non-Gaussian statistical structure of ground truth materials (details in \autoref{sec:non-gaussian}), and using a more complex prior, e.g., a flow-model-transformed Gaussian~\citep{yang2019pointflow}, might resolve this issue.

\subsection{Property optimization}

\begin{table}[t]
\caption{Property optimization performance.}
\scriptsize
\label{table:property-optimization}
\begin{center}
\begin{tabular}{l|lllllllll}
\multirow{2}{*}{\bf Method}  &\multicolumn{3}{c}{\bf Perov-5} & \multicolumn{3}{c}{\bf Carbon-24} & \multicolumn{3}{c}{\bf MP-20}\\
 & \multicolumn{1}{c}{SR5} & \multicolumn{1}{c}{SR10} & \multicolumn{1}{c}{SR15} & \multicolumn{1}{c}{SR5} & \multicolumn{1}{c}{SR10} & \multicolumn{1}{c}{SR15} & \multicolumn{1}{c}{SR5} & \multicolumn{1}{c}{SR10} & \multicolumn{1}{c}{SR15} \\
 \hline 
FTCP & 0.06 & 0.11 & 0.16 & 0.0 & 0.0 & 0.0 & 0.02 & 0.04 & 0.05 \\
Cond-DFC-VAE & \REVISION{\textbf{0.55}} & \REVISION{0.64} & \REVISION{0.69} & -- & -- & -- & -- & -- & -- \\
CDVAE & 0.52 & \textbf{0.65} & \textbf{0.79} & 0.0 & \textbf{0.06} & \textbf{0.06} & \REVISION{\textbf{0.78}} & \REVISION{\textbf{0.86}} & \REVISION{\textbf{0.90}} \\
\end{tabular}
\end{center}
\end{table}

\textbf{Setup.} The third task is to generate materials that optimize a specific property. Following \cite{jin2018junction}, we jointly train a property predictor $F$ parameterized by an MLP to predict properties of training materials from latent $\vz$. To optimize properties, we start with the latent representations of testing materials and apply gradient ascent in the latent space to improve the predicted property $F(\cdot)$. After applying 5000 gradient steps with step sizes of $1 \times 10^{-3}$, 10 materials are decoded from the latent trajectories every 500 steps. We use an independently trained property predictor to select the best one from the 10 decoded materials. \REVISION{Cond-DFC-VAE is a conditional VAE so we directly condition on the target property, sample 10 materials, and select the best one using the property predictor. For all methods, we generate 100 materials following the protocol above.} We use the independent property predictor to predict the properties for evaluation. We report the success rate (SR) as the percentage of materials achieving 5, 10, and 15 percentiles of the target property distribution. Our task is to minimize formation energy per atom for all 3 datasets.

\textbf{Results.} The performance is shown in \autoref{table:property-optimization}. \REVISION{We significantly outperform FTCP, while having a similar performance as Cond-DFC-VAE in Perov-5 (Cond-DFC-VAE cannot work for Carbon-24 and MP-20).} Both G-SchNet \REVISION{and P-G-SchNet} are incapable of property optimization \footnote{Very recently the authors published an improved version for conditional generation \citep{gebauer2021inverse} but the code has not been released yet.}. We note that all models perform poorly on the Carbon-24 dataset, which might be explained by the complex and diverse 3D structures of carbon.


\section{Conclusions and Outlook}

We have introduced a Crystal Diffusion Variational Autoencoder (CDVAE) to generate the periodic structure of stable materials and demonstrated that it significantly outperforms past methods on the tasks of reconstruction, generation, and property optimization. We note that the last two tasks are far more important for material design than reconstruction because they can be directly used to generate new materials whose properties can then be verified by QM simulations and experiments. We believe CDVAE opens up exciting opportunities for the inverse design of materials for various important applications. Meanwhile, our model is just a first step towards the grand challenge of material design. We provide our datasets and evaluation metrics to the broader machine learning community to collectively develop better methods for the task of material generation.

\section*{Reproducibility Statement}

We have made the following efforts to ensure reproducibility: 1) We provide our code at \url{https://github.com/txie-93/cdvae}; 2)We provide our data and corresponding train/validation/test splits at \url{https://github.com/txie-93/cdvae/tree/main/data}; 3) We provide details on experimental configurations in \autoref{sec:experimental-details}.


\section*{Acknowledgments}
We thank Peter Mikhael, Jason Yim, Rachel Wu, Bracha Laufer, Gabriele Corso, Felix Faltings, Bowen Jing, and the rest of the RB and TJ group members for their helpful comments and suggestions. The authors gratefully thank DARPA (HR00111920025), the consortium Machine Learning for Pharmaceutical Discovery and Synthesis (mlpds.mit.edu), and MIT-GIST collaboration for support.

\bibliography{ref}

\begin{thebibliography}{69}
\providecommand{\natexlab}[1]{#1}
\providecommand{\url}[1]{\texttt{#1}}
\expandafter\ifx\csname urlstyle\endcsname\relax
  \providecommand{\doi}[1]{doi: #1}\else
  \providecommand{\doi}{doi: \begingroup \urlstyle{rm}\Url}\fi

\bibitem[Belsky et~al.(2002)Belsky, Hellenbrandt, Karen, and
  Luksch]{belsky2002new}
Alec Belsky, Mariette Hellenbrandt, Vicky~Lynn Karen, and Peter Luksch.
\newblock New developments in the inorganic crystal structure database (icsd):
  accessibility in support of materials research and design.
\newblock \emph{Acta Crystallographica Section B: Structural Science},
  58\penalty0 (3):\penalty0 364--369, 2002.

\bibitem[Butler et~al.(2018)Butler, Davies, Cartwright, Isayev, and
  Walsh]{butler2018machine}
Keith~T Butler, Daniel~W Davies, Hugh Cartwright, Olexandr Isayev, and Aron
  Walsh.
\newblock Machine learning for molecular and materials science.
\newblock \emph{Nature}, 559\penalty0 (7715):\penalty0 547--555, 2018.

\bibitem[Cai et~al.(2020)Cai, Yang, Averbuch-Elor, Hao, Belongie, Snavely, and
  Hariharan]{cai2020learning}
Ruojin Cai, Guandao Yang, Hadar Averbuch-Elor, Zekun Hao, Serge Belongie, Noah
  Snavely, and Bharath Hariharan.
\newblock Learning gradient fields for shape generation.
\newblock In \emph{Computer Vision--ECCV 2020: 16th European Conference,
  Glasgow, UK, August 23--28, 2020, Proceedings, Part III 16}, pp.\  364--381.
  Springer, 2020.

\bibitem[Castelli et~al.(2012{\natexlab{a}})Castelli, Landis, Thygesen, Dahl,
  Chorkendorff, Jaramillo, and Jacobsen]{castelli2012new}
Ivano~E Castelli, David~D Landis, Kristian~S Thygesen, S{\o}ren Dahl,
  Ib~Chorkendorff, Thomas~F Jaramillo, and Karsten~W Jacobsen.
\newblock New cubic perovskites for one-and two-photon water splitting using
  the computational materials repository.
\newblock \emph{Energy \& Environmental Science}, 5\penalty0 (10):\penalty0
  9034--9043, 2012{\natexlab{a}}.

\bibitem[Castelli et~al.(2012{\natexlab{b}})Castelli, Olsen, Datta, Landis,
  Dahl, Thygesen, and Jacobsen]{castelli2012computational}
Ivano~E Castelli, Thomas Olsen, Soumendu Datta, David~D Landis, S{\o}ren Dahl,
  Kristian~S Thygesen, and Karsten~W Jacobsen.
\newblock Computational screening of perovskite metal oxides for optimal solar
  light capture.
\newblock \emph{Energy \& Environmental Science}, 5\penalty0 (2):\penalty0
  5814--5819, 2012{\natexlab{b}}.

\bibitem[Chanussot et~al.(2021)Chanussot, Das, Goyal, Lavril, Shuaibi, Riviere,
  Tran, Heras-Domingo, Ho, Hu, et~al.]{chanussot2021open}
Lowik Chanussot, Abhishek Das, Siddharth Goyal, Thibaut Lavril, Muhammed
  Shuaibi, Morgane Riviere, Kevin Tran, Javier Heras-Domingo, Caleb Ho, Weihua
  Hu, et~al.
\newblock Open catalyst 2020 (oc20) dataset and community challenges.
\newblock \emph{ACS Catalysis}, 11\penalty0 (10):\penalty0 6059--6072, 2021.

\bibitem[Chen \& Ong(2022)Chen and Ong]{chen2022universal}
Chi Chen and Shyue~Ping Ong.
\newblock A universal graph deep learning interatomic potential for the
  periodic table.
\newblock \emph{arXiv preprint arXiv:2202.02450}, 2022.

\bibitem[Chen et~al.(2019)Chen, Ye, Zuo, Zheng, and Ong]{chen2019graph}
Chi Chen, Weike Ye, Yunxing Zuo, Chen Zheng, and Shyue~Ping Ong.
\newblock Graph networks as a universal machine learning framework for
  molecules and crystals.
\newblock \emph{Chemistry of Materials}, 31\penalty0 (9):\penalty0 3564--3572,
  2019.

\bibitem[Court et~al.(2020)Court, Yildirim, Jain, and Cole]{court20203}
Callum~J Court, Batuhan Yildirim, Apoorv Jain, and Jacqueline~M Cole.
\newblock 3-d inorganic crystal structure generation and property prediction
  via representation learning.
\newblock \emph{Journal of chemical information and modeling}, 60\penalty0
  (10):\penalty0 4518--4535, 2020.

\bibitem[Dan et~al.(2020)Dan, Zhao, Li, Li, Hu, and Hu]{dan2020generative}
Yabo Dan, Yong Zhao, Xiang Li, Shaobo Li, Ming Hu, and Jianjun Hu.
\newblock Generative adversarial networks (gan) based efficient sampling of
  chemical composition space for inverse design of inorganic materials.
\newblock \emph{npj Computational Materials}, 6\penalty0 (1):\penalty0 1--7,
  2020.

\bibitem[Davies et~al.(2019)Davies, Butler, Jackson, Skelton, Morita, and
  Walsh]{davies2019smact}
Daniel~W Davies, Keith~T Butler, Adam~J Jackson, Jonathan~M Skelton, Kazuki
  Morita, and Aron Walsh.
\newblock Smact: Semiconducting materials by analogy and chemical theory.
\newblock \emph{Journal of Open Source Software}, 4\penalty0 (38):\penalty0
  1361, 2019.

\bibitem[Deringer et~al.(2018)Deringer, Pickard, and
  Cs{\'a}nyi]{deringer2018data}
Volker~L Deringer, Chris~J Pickard, and G{\'a}bor Cs{\'a}nyi.
\newblock Data-driven learning of total and local energies in elemental boron.
\newblock \emph{Physical review letters}, 120\penalty0 (15):\penalty0 156001,
  2018.

\bibitem[Dhariwal \& Nichol(2021)Dhariwal and Nichol]{dhariwal2021diffusion}
Prafulla Dhariwal and Alex Nichol.
\newblock Diffusion models beat gans on image synthesis.
\newblock \emph{arXiv preprint arXiv:2105.05233}, 2021.

\bibitem[Fuchs et~al.(2020)Fuchs, Worrall, Fischer, and
  Welling]{nips/FuchsW0W20}
Fabian Fuchs, Daniel~E. Worrall, Volker Fischer, and Max Welling.
\newblock Se(3)-transformers: 3d roto-translation equivariant attention
  networks.
\newblock In Hugo Larochelle, Marc'Aurelio Ranzato, Raia Hadsell,
  Maria{-}Florina Balcan, and Hsuan{-}Tien Lin (eds.), \emph{Advances in Neural
  Information Processing Systems 33: Annual Conference on Neural Information
  Processing Systems 2020, NeurIPS 2020, December 6-12, 2020, virtual}, 2020.
\newblock URL
  \url{https://proceedings.neurips.cc/paper/2020/hash/15231a7ce4ba789d13b722cc5c955834-Abstract.html}.

\bibitem[Ganea et~al.(2021)Ganea, Pattanaik, Coley, Barzilay, Jensen, Green,
  and Jaakkola]{ganea2021geomol}
Octavian-Eugen Ganea, Lagnajit Pattanaik, Connor~W Coley, Regina Barzilay,
  Klavs~F Jensen, William~H Green, and Tommi~S Jaakkola.
\newblock Geomol: Torsional geometric generation of molecular 3d conformer
  ensembles.
\newblock \emph{arXiv preprint arXiv:2106.07802}, 2021.

\bibitem[Gebauer et~al.(2019)Gebauer, Gastegger, and
  Sch\"{u}tt]{gebauer2019symmetry}
Niklas Gebauer, Michael Gastegger, and Kristof Sch\"{u}tt.
\newblock Symmetry-adapted generation of 3d point sets for the targeted
  discovery of molecules.
\newblock In H.~Wallach, H.~Larochelle, A.~Beygelzimer, F.~d\textquotesingle
  Alch\'{e}-Buc, E.~Fox, and R.~Garnett (eds.), \emph{Advances in Neural
  Information Processing Systems}, volume~32. Curran Associates, Inc., 2019.

\bibitem[Gebauer et~al.(2021)Gebauer, Gastegger, Hessmann, M{\"u}ller, and
  Sch{\"u}tt]{gebauer2021inverse}
Niklas~WA Gebauer, Michael Gastegger, Stefaan~SP Hessmann, Klaus-Robert
  M{\"u}ller, and Kristof~T Sch{\"u}tt.
\newblock Inverse design of 3d molecular structures with conditional generative
  neural networks.
\newblock \emph{arXiv preprint arXiv:2109.04824}, 2021.

\bibitem[Glass et~al.(2006)Glass, Oganov, and Hansen]{glass2006uspex}
Colin~W Glass, Artem~R Oganov, and Nikolaus Hansen.
\newblock Uspex—evolutionary crystal structure prediction.
\newblock \emph{Computer physics communications}, 175\penalty0
  (11-12):\penalty0 713--720, 2006.

\bibitem[Godwin et~al.(2021)Godwin, Schaarschmidt, Gaunt, Sanchez-Gonzalez,
  Rubanova, Veli{\v{c}}kovi{\'c}, Kirkpatrick, and Battaglia]{godwin2021very}
Jonathan Godwin, Michael Schaarschmidt, Alexander Gaunt, Alvaro
  Sanchez-Gonzalez, Yulia Rubanova, Petar Veli{\v{c}}kovi{\'c}, James
  Kirkpatrick, and Peter Battaglia.
\newblock Very deep graph neural networks via noise regularisation.
\newblock \emph{arXiv preprint arXiv:2106.07971}, 2021.

\bibitem[Grosse-Kunstleve et~al.(2004)Grosse-Kunstleve, Sauter, and
  Adams]{grosse2004numerically}
Ralf~W Grosse-Kunstleve, Nicholas~K Sauter, and Paul~D Adams.
\newblock Numerically stable algorithms for the computation of reduced unit
  cells.
\newblock \emph{Acta Crystallographica Section A: Foundations of
  Crystallography}, 60\penalty0 (1):\penalty0 1--6, 2004.

\bibitem[Hautier et~al.(2011)Hautier, Fischer, Ehrlacher, Jain, and
  Ceder]{hautier2011data}
Geoffroy Hautier, Chris Fischer, Virginie Ehrlacher, Anubhav Jain, and Gerbrand
  Ceder.
\newblock Data mined ionic substitutions for the discovery of new compounds.
\newblock \emph{Inorganic chemistry}, 50\penalty0 (2):\penalty0 656--663, 2011.

\bibitem[Ho et~al.(2020)Ho, Jain, and Abbeel]{nips/HoJA20}
Jonathan Ho, Ajay Jain, and Pieter Abbeel.
\newblock Denoising diffusion probabilistic models.
\newblock In Hugo Larochelle, Marc'Aurelio Ranzato, Raia Hadsell,
  Maria{-}Florina Balcan, and Hsuan{-}Tien Lin (eds.), \emph{Advances in Neural
  Information Processing Systems 33: Annual Conference on Neural Information
  Processing Systems 2020, NeurIPS 2020, December 6-12, 2020, virtual}, 2020.
\newblock URL
  \url{https://proceedings.neurips.cc/paper/2020/hash/4c5bcfec8584af0d967f1ab10179ca4b-Abstract.html}.

\bibitem[Hoffmann et~al.(2019)Hoffmann, Maestrati, Sawada, Tang, Sellier, and
  Bengio]{hoffmann2019data}
Jordan Hoffmann, Louis Maestrati, Yoshihide Sawada, Jian Tang, Jean~Michel
  Sellier, and Yoshua Bengio.
\newblock Data-driven approach to encoding and decoding 3-d crystal structures.
\newblock \emph{arXiv preprint arXiv:1909.00949}, 2019.

\bibitem[Hu et~al.(2021)Hu, Yang, Dong, Li, Li, Li, and
  Siriwardane]{hu2021contact}
Jianjun Hu, Wenhui Yang, Rongzhi Dong, Yuxin Li, Xiang Li, Shaobo Li, and
  Edirisuriya~MD Siriwardane.
\newblock Contact map based crystal structure prediction using global
  optimization.
\newblock \emph{CrystEngComm}, 23\penalty0 (8):\penalty0 1765--1776, 2021.

\bibitem[Irwin \& Shoichet(2005)Irwin and Shoichet]{irwin2005zinc}
John~J Irwin and Brian~K Shoichet.
\newblock Zinc- a free database of commercially available compounds for virtual
  screening.
\newblock \emph{Journal of chemical information and modeling}, 45\penalty0
  (1):\penalty0 177--182, 2005.

\bibitem[Jain et~al.(2013)Jain, Ong, Hautier, Chen, Richards, Dacek, Cholia,
  Gunter, Skinner, Ceder, et~al.]{jain2013commentary}
Anubhav Jain, Shyue~Ping Ong, Geoffroy Hautier, Wei Chen, William~Davidson
  Richards, Stephen Dacek, Shreyas Cholia, Dan Gunter, David Skinner, Gerbrand
  Ceder, et~al.
\newblock Commentary: The materials project: A materials genome approach to
  accelerating materials innovation.
\newblock \emph{APL materials}, 1\penalty0 (1):\penalty0 011002, 2013.

\bibitem[Jin et~al.(2018)Jin, Barzilay, and Jaakkola]{jin2018junction}
Wengong Jin, Regina Barzilay, and Tommi Jaakkola.
\newblock Junction tree variational autoencoder for molecular graph generation.
\newblock In \emph{International conference on machine learning}, pp.\
  2323--2332. PMLR, 2018.

\bibitem[Kim et~al.(2020)Kim, Noh, Gu, Aspuru-Guzik, and
  Jung]{kim2020generative}
Sungwon Kim, Juhwan Noh, Geun~Ho Gu, Alan Aspuru-Guzik, and Yousung Jung.
\newblock Generative adversarial networks for crystal structure prediction.
\newblock \emph{ACS central science}, 6\penalty0 (8):\penalty0 1412--1420,
  2020.

\bibitem[Kingma \& Welling(2014)Kingma and Welling]{corr/KingmaW13}
Diederik~P. Kingma and Max Welling.
\newblock Auto-encoding variational bayes.
\newblock In Yoshua Bengio and Yann LeCun (eds.), \emph{2nd International
  Conference on Learning Representations, {ICLR} 2014, Banff, AB, Canada, April
  14-16, 2014, Conference Track Proceedings}, 2014.
\newblock URL \url{http://arxiv.org/abs/1312.6114}.

\bibitem[Klicpera et~al.(2020{\natexlab{a}})Klicpera, Giri, Margraf, and
  G{\"u}nnemann]{klicpera2020fast}
Johannes Klicpera, Shankari Giri, Johannes~T Margraf, and Stephan
  G{\"u}nnemann.
\newblock Fast and uncertainty-aware directional message passing for
  non-equilibrium molecules.
\newblock \emph{arXiv preprint arXiv:2011.14115}, 2020{\natexlab{a}}.

\bibitem[Klicpera et~al.(2020{\natexlab{b}})Klicpera, Gro{\ss}, and
  G{\"{u}}nnemann]{iclr/KlicperaGG20}
Johannes Klicpera, Janek Gro{\ss}, and Stephan G{\"{u}}nnemann.
\newblock Directional message passing for molecular graphs.
\newblock In \emph{8th International Conference on Learning Representations,
  {ICLR} 2020, Addis Ababa, Ethiopia, April 26-30, 2020}. OpenReview.net,
  2020{\natexlab{b}}.
\newblock URL \url{https://openreview.net/forum?id=B1eWbxStPH}.

\bibitem[Klicpera et~al.(2021)Klicpera, Becker, and
  G{\"u}nnemann]{klicpera2021gemnet}
Johannes Klicpera, Florian Becker, and Stephan G{\"u}nnemann.
\newblock Gemnet: Universal directional graph neural networks for molecules.
\newblock \emph{arXiv preprint arXiv:2106.08903}, 2021.

\bibitem[Kong \& Ping(2021)Kong and Ping]{kong2021fast}
Zhifeng Kong and Wei Ping.
\newblock On fast sampling of diffusion probabilistic models.
\newblock \emph{arXiv preprint arXiv:2106.00132}, 2021.

\bibitem[Liu et~al.(2021)Liu, Yan, Oztekin, and Ji]{liu2021graphebm}
Meng Liu, Keqiang Yan, Bora Oztekin, and Shuiwang Ji.
\newblock Graphebm: Molecular graph generation with energy-based models.
\newblock \emph{arXiv preprint arXiv:2102.00546}, 2021.

\bibitem[Long et~al.(2021)Long, Fortunato, Opahle, Zhang, Samathrakis, Shen,
  Gutfleisch, and Zhang]{long2021constrained}
Teng Long, Nuno~M Fortunato, Ingo Opahle, Yixuan Zhang, Ilias Samathrakis, Chen
  Shen, Oliver Gutfleisch, and Hongbin Zhang.
\newblock Constrained crystals deep convolutional generative adversarial
  network for the inverse design of crystal structures.
\newblock \emph{npj Computational Materials}, 7\penalty0 (1):\penalty0 1--7,
  2021.

\bibitem[Luo \& Hu(2021)Luo and Hu]{luo2021diffusion}
Shitong Luo and Wei Hu.
\newblock Diffusion probabilistic models for 3d point cloud generation.
\newblock In \emph{Proceedings of the IEEE/CVF Conference on Computer Vision
  and Pattern Recognition}, pp.\  2837--2845, 2021.

\bibitem[Luo et~al.(2021)Luo, Yan, and Ji]{luo2021graphdf}
Youzhi Luo, Keqiang Yan, and Shuiwang Ji.
\newblock Graphdf: A discrete flow model for molecular graph generation.
\newblock \emph{arXiv preprint arXiv:2102.01189}, 2021.

\bibitem[Nichol \& Dhariwal(2021)Nichol and Dhariwal]{nichol2021improved}
Alexander~Quinn Nichol and Prafulla Dhariwal.
\newblock Improved denoising diffusion probabilistic models.
\newblock In \emph{International Conference on Machine Learning}, pp.\
  8162--8171. PMLR, 2021.

\bibitem[Noh et~al.(2019)Noh, Kim, Stein, Sanchez-Lengeling, Gregoire,
  Aspuru-Guzik, and Jung]{noh2019inverse}
Juhwan Noh, Jaehoon Kim, Helge~S Stein, Benjamin Sanchez-Lengeling, John~M
  Gregoire, Alan Aspuru-Guzik, and Yousung Jung.
\newblock Inverse design of solid-state materials via a continuous
  representation.
\newblock \emph{Matter}, 1\penalty0 (5):\penalty0 1370--1384, 2019.

\bibitem[Oganov et~al.(2019)Oganov, Pickard, Zhu, and
  Needs]{oganov2019structure}
Artem~R Oganov, Chris~J Pickard, Qiang Zhu, and Richard~J Needs.
\newblock Structure prediction drives materials discovery.
\newblock \emph{Nature Reviews Materials}, 4\penalty0 (5):\penalty0 331--348,
  2019.

\bibitem[O'Keeffe \& Hyde(1980)O'Keeffe and Hyde]{10.2307/36648}
M.~O'Keeffe and B.~G. Hyde.
\newblock Plane nets in crystal chemistry.
\newblock \emph{Philosophical Transactions of the Royal Society of London.
  Series A, Mathematical and Physical Sciences}, 295\penalty0 (1417):\penalty0
  553--618, 1980.
\newblock ISSN 00804614.
\newblock URL \url{http://www.jstor.org/stable/36648}.

\bibitem[Ong et~al.(2013)Ong, Richards, Jain, Hautier, Kocher, Cholia, Gunter,
  Chevrier, Persson, and Ceder]{ong2013python}
Shyue~Ping Ong, William~Davidson Richards, Anubhav Jain, Geoffroy Hautier,
  Michael Kocher, Shreyas Cholia, Dan Gunter, Vincent~L Chevrier, Kristin~A
  Persson, and Gerbrand Ceder.
\newblock Python materials genomics (pymatgen): A robust, open-source python
  library for materials analysis.
\newblock \emph{Computational Materials Science}, 68:\penalty0 314--319, 2013.

\bibitem[Pan et~al.(2021)Pan, Ganose, Horton, Aykol, Persson, Zimmermann, and
  Jain]{pan2021benchmarking}
Hillary Pan, Alex~M Ganose, Matthew Horton, Muratahan Aykol, Kristin~A Persson,
  Nils~ER Zimmermann, and Anubhav Jain.
\newblock Benchmarking coordination number prediction algorithms on inorganic
  crystal structures.
\newblock \emph{Inorganic chemistry}, 60\penalty0 (3):\penalty0 1590--1603,
  2021.

\bibitem[Pathak et~al.(2020)Pathak, Juneja, Varma, Ehara, and
  Priyakumar]{pathak2020deep}
Yashaswi Pathak, Karandeep~Singh Juneja, Girish Varma, Masahiro Ehara, and
  U~Deva Priyakumar.
\newblock Deep learning enabled inorganic material generator.
\newblock \emph{Physical Chemistry Chemical Physics}, 22\penalty0
  (46):\penalty0 26935--26943, 2020.

\bibitem[Pickard(2020)]{carbon2020data}
Chris~J. Pickard.
\newblock Airss data for carbon at 10gpa and the c+n+h+o system at 1gpa, 2020.
\newblock URL \url{https://archive.materialscloud.org/record/2020.0026/v1}.

\bibitem[Pickard \& Needs(2006)Pickard and Needs]{pickard2006high}
Chris~J Pickard and RJ~Needs.
\newblock High-pressure phases of silane.
\newblock \emph{Physical review letters}, 97\penalty0 (4):\penalty0 045504,
  2006.

\bibitem[Pickard \& Needs(2011)Pickard and Needs]{pickard2011ab}
Chris~J Pickard and RJ~Needs.
\newblock Ab initio random structure searching.
\newblock \emph{Journal of Physics: Condensed Matter}, 23\penalty0
  (5):\penalty0 053201, 2011.

\bibitem[Ren et~al.(2020)Ren, Noh, Tian, Oviedo, Xing, Liang, Aberle, Liu, Li,
  Jayavelu, et~al.]{ren2020inverse}
Zekun Ren, Juhwan Noh, Siyu Tian, Felipe Oviedo, Guangzong Xing, Qiaohao Liang,
  Armin Aberle, Yi~Liu, Qianxiao Li, Senthilnath Jayavelu, et~al.
\newblock Inverse design of crystals using generalized invertible
  crystallographic representation.
\newblock \emph{arXiv preprint arXiv:2005.07609}, 2020.

\bibitem[Salimans \& Ho(2022)Salimans and Ho]{salimans2022progressive}
Tim Salimans and Jonathan Ho.
\newblock Progressive distillation for fast sampling of diffusion models.
\newblock \emph{arXiv preprint arXiv:2202.00512}, 2022.

\bibitem[Satorras et~al.(2021)Satorras, Hoogeboom, Fuchs, Posner, and
  Welling]{satorras2021n}
Victor~Garcia Satorras, Emiel Hoogeboom, Fabian~B Fuchs, Ingmar Posner, and Max
  Welling.
\newblock E (n) equivariant normalizing flows for molecule generation in 3d.
\newblock \emph{arXiv preprint arXiv:2105.09016}, 2021.

\bibitem[Sawada et~al.(2019)Sawada, Morikawa, and Fujii]{sawada2019study}
Yoshihide Sawada, Koji Morikawa, and Mikiya Fujii.
\newblock Study of deep generative models for inorganic chemical compositions.
\newblock \emph{arXiv preprint arXiv:1910.11499}, 2019.

\bibitem[Sch{\"u}tt et~al.(2018)Sch{\"u}tt, Sauceda, Kindermans, Tkatchenko,
  and M{\"u}ller]{schutt2018schnet}
Kristof~T Sch{\"u}tt, Huziel~E Sauceda, P-J Kindermans, Alexandre Tkatchenko,
  and K-R M{\"u}ller.
\newblock Schnet--a deep learning architecture for molecules and materials.
\newblock \emph{The Journal of Chemical Physics}, 148\penalty0 (24):\penalty0
  241722, 2018.

\bibitem[Shi et~al.(2020)Shi, Xu, Zhu, Zhang, Zhang, and Tang]{shi2020graphaf}
Chence Shi, Minkai Xu, Zhaocheng Zhu, Weinan Zhang, Ming Zhang, and Jian Tang.
\newblock Graphaf: a flow-based autoregressive model for molecular graph
  generation.
\newblock \emph{arXiv preprint arXiv:2001.09382}, 2020.

\bibitem[Shi et~al.(2021)Shi, Luo, Xu, and Tang]{icml/ShiLXT21}
Chence Shi, Shitong Luo, Minkai Xu, and Jian Tang.
\newblock Learning gradient fields for molecular conformation generation.
\newblock In Marina Meila and Tong Zhang (eds.), \emph{Proceedings of the 38th
  International Conference on Machine Learning, {ICML} 2021, 18-24 July 2021,
  Virtual Event}, volume 139 of \emph{Proceedings of Machine Learning
  Research}, pp.\  9558--9568. {PMLR}, 2021.
\newblock URL \url{http://proceedings.mlr.press/v139/shi21b.html}.

\bibitem[Shuaibi et~al.(2021)Shuaibi, Kolluru, Das, Grover, Sriram, Ulissi, and
  Zitnick]{shuaibi2021rotation}
Muhammed Shuaibi, Adeesh Kolluru, Abhishek Das, Aditya Grover, Anuroop Sriram,
  Zachary Ulissi, and C~Lawrence Zitnick.
\newblock Rotation invariant graph neural networks using spin convolutions.
\newblock \emph{arXiv preprint arXiv:2106.09575}, 2021.

\bibitem[Sohl-Dickstein et~al.(2015)Sohl-Dickstein, Weiss, Maheswaranathan, and
  Ganguli]{sohl2015deep}
Jascha Sohl-Dickstein, Eric Weiss, Niru Maheswaranathan, and Surya Ganguli.
\newblock Deep unsupervised learning using nonequilibrium thermodynamics.
\newblock In \emph{International Conference on Machine Learning}, pp.\
  2256--2265. PMLR, 2015.

\bibitem[Song \& Ermon(2019)Song and Ermon]{nips/SongE19}
Yang Song and Stefano Ermon.
\newblock Generative modeling by estimating gradients of the data distribution.
\newblock In Hanna~M. Wallach, Hugo Larochelle, Alina Beygelzimer, Florence
  d'Alch{\'{e}}{-}Buc, Emily~B. Fox, and Roman Garnett (eds.), \emph{Advances
  in Neural Information Processing Systems 32: Annual Conference on Neural
  Information Processing Systems 2019, NeurIPS 2019, December 8-14, 2019,
  Vancouver, BC, Canada}, pp.\  11895--11907, 2019.
\newblock URL
  \url{https://proceedings.neurips.cc/paper/2019/hash/3001ef257407d5a371a96dcd947c7d93-Abstract.html}.

\bibitem[Thomas et~al.(2018)Thomas, Smidt, Kearnes, Yang, Li, Kohlhoff, and
  Riley]{thomas2018tensor}
Nathaniel Thomas, Tess Smidt, Steven Kearnes, Lusann Yang, Li~Li, Kai Kohlhoff,
  and Patrick Riley.
\newblock Tensor field networks: Rotation-and translation-equivariant neural
  networks for 3d point clouds.
\newblock \emph{arXiv preprint arXiv:1802.08219}, 2018.

\bibitem[Wang et~al.(2012)Wang, Lv, Zhu, and Ma]{wang2012calypso}
Yanchao Wang, Jian Lv, Li~Zhu, and Yanming Ma.
\newblock Calypso: A method for crystal structure prediction.
\newblock \emph{Computer Physics Communications}, 183\penalty0 (10):\penalty0
  2063--2070, 2012.

\bibitem[Ward et~al.(2016)Ward, Agrawal, Choudhary, and
  Wolverton]{ward2016general}
Logan Ward, Ankit Agrawal, Alok Choudhary, and Christopher Wolverton.
\newblock A general-purpose machine learning framework for predicting
  properties of inorganic materials.
\newblock \emph{npj Computational Materials}, 2\penalty0 (1):\penalty0 1--7,
  2016.

\bibitem[Wells et~al.(1977)]{wells1977three}
Alexander~Frank Wells et~al.
\newblock \emph{Three dimensional nets and polyhedra}.
\newblock Wiley, 1977.

\bibitem[Wu et~al.(2021)Wu, Shen, Lan, Bian, and Huang]{wu2021se}
Jiaxiang Wu, Tao Shen, Haidong Lan, Yatao Bian, and Junzhou Huang.
\newblock Se (3)-equivariant energy-based models for end-to-end protein
  folding.
\newblock \emph{bioRxiv}, 2021.

\bibitem[Xie \& Grossman(2018)Xie and Grossman]{xie2018crystal}
Tian Xie and Jeffrey~C Grossman.
\newblock Crystal graph convolutional neural networks for an accurate and
  interpretable prediction of material properties.
\newblock \emph{Physical review letters}, 120\penalty0 (14):\penalty0 145301,
  2018.

\bibitem[Xu et~al.(2021{\natexlab{a}})Xu, Luo, Bengio, Peng, and
  Tang]{xu2021learning}
Minkai Xu, Shitong Luo, Yoshua Bengio, Jian Peng, and Jian Tang.
\newblock Learning neural generative dynamics for molecular conformation
  generation.
\newblock In \emph{International Conference on Learning Representations},
  2021{\natexlab{a}}.
\newblock URL \url{https://openreview.net/forum?id=pAbm1qfheGk}.

\bibitem[Xu et~al.(2021{\natexlab{b}})Xu, Yu, Song, Shi, Ermon, and
  Tang]{xu2021geodiff}
Minkai Xu, Lantao Yu, Yang Song, Chence Shi, Stefano Ermon, and Jian Tang.
\newblock Geodiff: A geometric diffusion model for molecular conformation
  generation.
\newblock In \emph{International Conference on Learning Representations},
  2021{\natexlab{b}}.

\bibitem[Yang et~al.(2019)Yang, Huang, Hao, Liu, Belongie, and
  Hariharan]{yang2019pointflow}
Guandao Yang, Xun Huang, Zekun Hao, Ming-Yu Liu, Serge Belongie, and Bharath
  Hariharan.
\newblock Pointflow: 3d point cloud generation with continuous normalizing
  flows.
\newblock In \emph{Proceedings of the IEEE/CVF International Conference on
  Computer Vision}, pp.\  4541--4550, 2019.

\bibitem[Yang et~al.(2021)Yang, Siriwardane, Dong, Li, and Hu]{yang2021crystal}
Wenhui Yang, Edirisuriya M~Dilanga Siriwardane, Rongzhi Dong, Yuxin Li, and
  Jianjun Hu.
\newblock Crystal structure prediction of materials with high symmetry using
  differential evolution.
\newblock \emph{arXiv preprint arXiv:2104.09764}, 2021.

\bibitem[Zhao et~al.(2021)Zhao, Al-Fahdi, Hu, Siriwardane, Song, Nasiri, and
  Hu]{zhao2021high}
Yong Zhao, Mohammed Al-Fahdi, Ming Hu, Edirisuriya Siriwardane, Yuqi Song,
  Alireza Nasiri, and Jianjun Hu.
\newblock High-throughput discovery of novel cubic crystal materials using deep
  generative neural networks.
\newblock \emph{arXiv preprint arXiv:2102.01880}, 2021.

\bibitem[Zimmermann \& Jain(2020)Zimmermann and Jain]{zimmermann2020local}
Nils~ER Zimmermann and Anubhav Jain.
\newblock Local structure order parameters and site fingerprints for
  quantification of coordination environment and crystal structure similarity.
\newblock \emph{RSC Advances}, 10\penalty0 (10):\penalty0 6063--6081, 2020.

\end{thebibliography}
\bibliographystyle{iclr2022_conference}

\newpage

\appendix

\section{Proof for the connection to a harmonic force field} \label{sec:proof}

We assume the loss in \autoref{eq:loss} can be minimized to zero when the noises are small, meaning that
\begin{equation}
    \vs_\mX (\tilde{\mA}, \tilde{\mX}, \mL | \vz) = \frac{\vd_{\mathrm{min}}(\mX, \tilde{\mX})}{\sigma_{\mX,j}}, \forall j > J,
\end{equation}
where $\sigma_{\mX,j} \in  \{ \sigma_{\mX,j} \}_{j=1}^L$ and any noise smaller than $\sigma_{\mX, J}$ is considered as small.

The force term in the Langevin dynamics $\alpha_j \vs_{\mX, t}$ can then be written as
\begin{align}
     \alpha_j \vs_\mX(\tilde{\mA}, \tilde{\mX}, \mL | \vz; \sigma_{\mA,j}, \sigma_{\mX,j}) &= \epsilon \cdot \sigma_{\mX,j}^2/\sigma_{\mX,L}^2 \cdot \vs_\mX (\tilde{\mA}, \tilde{\mX}, \mL | \vz) / \sigma_{\mX,j} \\
     &= \epsilon \cdot \frac{\sigma_{\mX,j}^2}{\sigma_{\mX,L}^2} \cdot \frac{\vd_{\mathrm{min}}(\mX, \tilde{\mX})}{\sigma_{\mX,j}^2}, \forall j > J \\
     &= - \frac{\epsilon}{\sigma_{\mX,L}^2} \vd_{\mathrm{min}}(\tilde{\mX}, \mX), \forall j > J
\end{align}

If we write $\epsilon / \sigma_{\mX,L}^2 = k$, then,
\begin{equation}
    \alpha_j \vs_\mX(\tilde{\mA}, \tilde{\mX}, \mL | \vz; \sigma_{\mA,j}, \sigma_{\mX,j}) = -k \vd_{\mathrm{min}}(\tilde{\mX}, \mX), \forall j > J
\end{equation}

If the noises are small enough that atoms do not cross the periodic boundaries, then we have $\vd_{\mathrm{min}}(\mX, \tilde{\mX}) = \mX - \tilde{\mX}$. Therefore,
\begin{equation}
    \alpha_j \vs_\mX(\tilde{\mA}, \tilde{\mX}, \mL | \vz; \sigma_{\mA,j}, \sigma_{\mX,j}) = -k (\tilde{\mX} - \mX), \forall j > J.
\end{equation}

\section{Implementation details} \label{sec:implementation}

\subsection{Prediction of lattice parameters} \label{sec:lattice-prediction}

 There are infinitely many different ways of choosing the lattice for the same material. We compute the Niggli reduced lattice \citep{grosse2004numerically} with pymatgen \citep{ong2013python}, which is a unique lattice for any given material. Since the lattice matrix $\mL$ is not rotation invariant, we instead predict the 6 lattice parameters, i.e. the lengths of the 3 lattice vectors and the angles between them. We normalize the lengths of lattice vectors with $\sqrt[3]{N}$, where $N$ is the number of atoms, to ensure that the lengths for materials of different sizes are at the same scale.
 
\subsection{Multi-graph construction}

For the encoder, we use CrystalNN \citep{pan2021benchmarking} to determine edges between atoms and build a multi-graph representation. For the decoder, since it inputs a noisy structure generated on the fly, the multi-graph must also be built on the fly for both training and generation, and CrystalNN is too slow for that purpose. We use a KNN algorithm that considers periodicity to build the decoder graph where $K=20$ in all of our experiments.

\subsection{GNN architecture}
 
We use DimeNet++ adapted for periodicity \citep{klicpera2020fast,iclr/KlicperaGG20} as the encoder, which is SE(3) invariant to the input structure. The decoder needs to output an vector per node that is SE(3) equivariant to the input structure. We use GemNet-dQ \citep{klicpera2021gemnet} as the decoder. We used implementations from the Open Catalysis Project (OCP) \citep{chanussot2021open}, but we reduced the size of hidden dimensions to 128 for faster training. The encoder has 2.2 million parameters and the decoder has 2.3 million parameters.

\section{Dataset curation} \label{app-sec:dataset}

\subsection{Perov-5}

Perovskite is a class of materials that share a similar structure and have the general chemical formula \ce{ABX_3}. The ideal perovskites have a cubic structure, where the site A atom sits at a corner position, the site B atom sits at a body centered position and site X atoms sit at face centered positions. Perovskite materials are known for their wide applications. We curate the Perov-5 dataset from an open database that was originally developed for water splitting  \citep{castelli2012new,castelli2012computational}.

All 18928 materials in the original database are included. In the database, A, B can be any non-radioactive metal and X can be one or several elements from O, N, S, and F. Note that there can be multiple different X atoms in the same material. All materials in Perov-5 are relaxed using density functional theory (DFT), and their relaxed structure can deviate significantly from the ideal structures. A significant portion of the materials are not thermodynamically stable, i.e., they will decompose to nearby phases and cannot be synthesized.

\subsection{Carbon-24}

Carbon-24 includes various carbon structures obtained via \textit{ab initio} random structure searching (AIRSS) \citep{pickard2006high,pickard2011ab} performed at 10 GPa. 

The original dataset includes 101529 carbon structures, and we selected the 10\% of the carbon structure with the lowest energy per atom to create Carbon-24. All 10153 structures in Carbon-24 are relaxed using DFT. The most stable structure is diamond at 10 GPa. All remaining structures are thermodynamically unstable but may be kinetically stable. Most of the structures cannot be synthesized.

\subsection{MP-20}

MP-20 includes almost all experimentally stable materials from the Materials Project \citep{jain2013commentary} with unit cells including at most 20 atoms. \REVISION{We only include materials that are originally from ICSD \citep{belsky2002new} to ensure the experimental stability,} and these materials represent the majority of experimentally known materials with at most 20 atoms in unit cells.

To ensure stability, we only select materials with energy above the hull smaller than 0.08 eV/atom and formation energy smaller than 2 eV/atom, following \cite{ren2020inverse}. Differing from \cite{ren2020inverse}, we do not constrain the number of unique elements per material. All materials in MP-20 are relaxed using DFT. Most materials are thermodynamcially stable and have been synthesized.

\section{Experiment details} \label{sec:experimental-details}

\subsection{Reasons for the unsuitability of some metrics for specific datasets} \label{sec:unsuitable-metrics}

In \autoref{table:generation}, property statistics are computed by comparing the earth mover's distance between the property distribution of generated materials and ground truth materials. So, they are not meaningful for ground truth data.

Materials in Perov-5 have the same structure, so it is not meaningful to require higher structure diversity.

Materials in Carbon-24 have the same composition (carbon), so it is not meaningful to require higher composition diversity. In addition, all models have $\sim$100\% composition validity, so it is not compared in the table.

\subsection{Composition validity checker}

We modified the charge neutrality checker from \texttt{SMACT} \citep{davies2019smact} because the original checker is not suitable for alloys. The checker is based on a list of possible charges for each element and it checks if the material can be charge neutral by enumerating all possible charge combinations. However, it does not consider that metal alloys can be mixed with almost any combination. As a result, for materials composed of all metal elements, we always assume the composition is valid in our validity checker.

For the ground truth materials in MP-20, the original checker gives a composition validity of $\sim$50\%, which significantly underestimates the validity of MP-20 materials (because most of them are experimentally synthesizable and thus valid). Our checker gives a composition validity of $\sim$90\%, which is far more reasonable. We note again that these checkers are all empirical and the only high-fidelity evaluation of material stability requires QM simulations.

\subsection{Non-Gaussian statistical structure of materials} \label{sec:non-gaussian}

The material datasets are usually biased towards certain material groups. For example, there are lots of lithium-containing materials in MP-20 because it started with battery research. We also find that our decoder tends to underfit the data distribution with a larger $\beta$ in \autoref{eq:total-loss}. We believe these observations indicate that the statistical structure of the ground truth materials are far from Gaussian. As a result, sampling from $\mathcal{N}(0, 1)$ may lead to out-of-distribution materials, which explains why our method tends to generate more elements per material than the ground truth.

\subsection{Hyperparameters and training details}

The total loss can be written as,
\begin{equation} \label{eq:total-loss}
    \mathcal{L} = \mathcal{L}_{\textsc{Agg}} + \mathcal{L}_{\textsc{Dec}} + \mathcal{L}_{\textsc{KL}} = \lambda_\vc \mathcal{L}_{\vc} + \lambda_\mL \mathcal{L}_{\mL} + \lambda_N \mathcal{L}_{N} + \lambda_\mX \mathcal{L}_{\mX} + \lambda_\mA \mathcal{L}_{\mA} + \beta \mathcal{L}_{\mathrm{KL}}.
\end{equation}

We aim to keep each loss term at a similar scale. For all three datasets, we use $\lambda_\vc = 1, \lambda_\mL = 10, \lambda_N = 1, \lambda_\mX = 10, \mathcal{L}_{\mA} = 1$.

We tune $\beta$ between $0.01, 0.03, 0.1$ for all three datasets and select the model with best validation loss. For Perov-5, MP-20, we use $\beta = 0.01$, and for Carbon-24, we use $\beta = 0.03$.

\REVISION{For the noise levels in $\{ \sigma_{\mA,j} \}_{j=1}^L, \{ \sigma_{\mX,j} \}_{j=1}^L$, we follow \cite{icml/ShiLXT21} and set $L = 50$. For all three datasets, we use $\sigma_{\mA, \mathrm{max}} = 5, \sigma_{\mA, \mathrm{min}} = 0.01, \sigma_{\mX, \mathrm{max}} = 10, \sigma_{\mX, \mathrm{min}} = 0.01$.}

During the training, we use an initial learning rate of 0.001 and reduce the learning rate by a factor of 0.6 if the validation loss does not improve after 30 epochs. The minimum learning rate is 0.0001.

During the generation, we use $\epsilon = 0.0001$ and run Langevin dynamics for 100 steps at each noise level.

\FloatBarrier
\section{Visualization of multiple reconstructed structures}

\begin{figure}[htp]
    \centering
    \includegraphics[width=\textwidth]{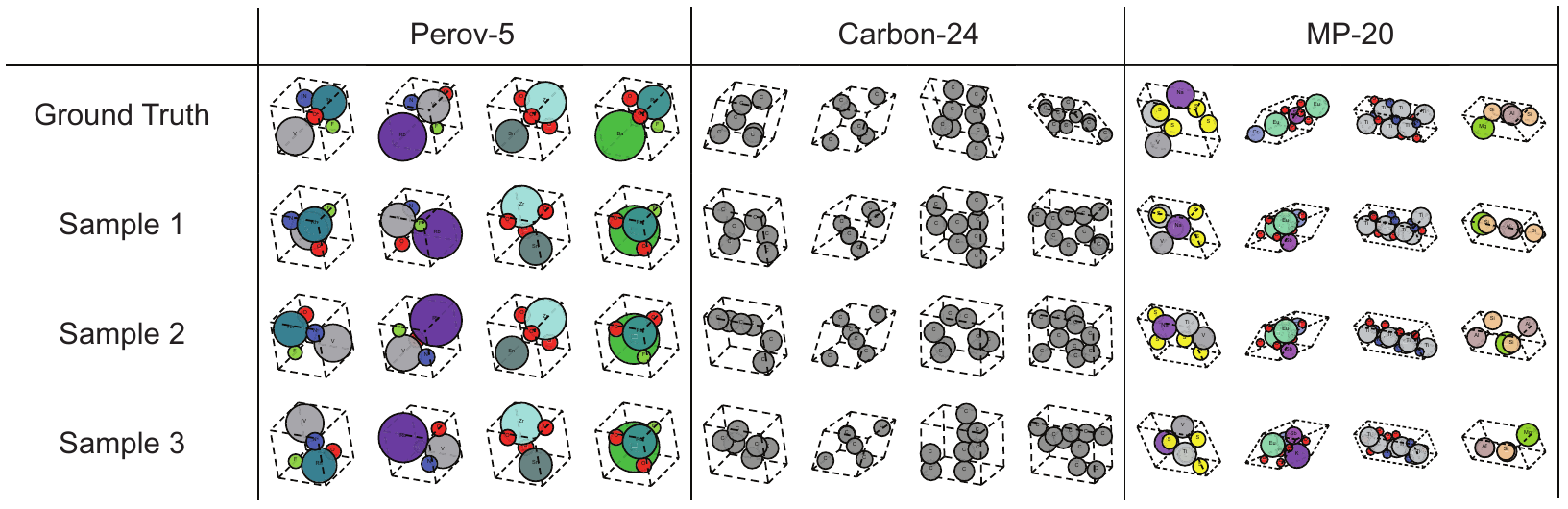}
    \caption{Different reconstructed structures from CDVAE from the same $\vz$, following 3 Langevin dynamics sampling with different random seeds.}
    \label{fig:sampling}
\end{figure}

\FloatBarrier
\section{Sampling speed for material generation}

We summarize the speed for generating 10,000 materials for all models in \autoref{table:sampling-speed}. FTCP is significantly faster, but the quality of generated materials is very poor as shown in \autoref{table:generation}. Cond-DFC-VAE is faster than our method in Perov-5, but has a lower quality than our method and only works for cubic systems. It is also unclear how it will perform on larger materials in Carbon-24 and MP-20, because the compute increases cubicily with the increased size of the density map. G-SchNet/P-G-SchNet have a comparable sampling time as our method, but have a lower quality. We also note that we did not optimize sampling speed in current work. It is possible to reduce sampling time by using fewer sampling steps without significantly influencing generation quality. There are also many recent works that aim to speed up the sampling process for diffusion models \citep{nichol2021improved,kong2021fast,salimans2022progressive}.

\begin{table}[h]
\caption{Time used for generating 10,000 materials on a single RTX 2080 Ti GPU.}
\label{table:sampling-speed}
\begin{center}
\begin{tabular}{l|ccccc}
 & FTCP & Cond-DFC-VAE & G-SchNet & P-G-SchNet & CDVAE \\
 \hline 
Perov-5 &  $<1$ min & 0.5 h & 2.0 h & 2.0 h & 3.1 h\\
Carbon-24 & $<1$ min & -- & 6.2 h & 6.3 h & 5.3 h \\
MP-20 & $<1$ min & -- & 6.3 h & 6.3 h & 5.8 h \\
\end{tabular}
\end{center}
\end{table}

\FloatBarrier
\section{Coverage metrics for material generation} \label{sec:coverage}

Inspired by \cite{xu2021learning,ganea2021geomol}, we define six metrics to compare two ensembles of materials: materials generated by a method $\{ \mM_k \}_{k \in [1..K]}$, and ground truth materials in test data $\{ \mM_l^* \}_{ \in [1..L]}$. 

We use the Euclidean distance of the CrystalNN fingerprint \citep{zimmermann2020local} and normalized Magpie fingerprint \citep{ward2016general} to define the structure distance and composition distance between generated and ground truth materials, respectively. They can be written as $D_{\mathrm{struc.}}(\mM_k, \mM_l^*)$ and $D_{\mathrm{comp.}}(\mM_k, \mM_l^*)$. We further define the thresholds for the structure and composition distance as $\delta_{\mathrm{struc.}}$ and $\delta_{\mathrm{comp.}}$, respectively.

Following the established classification metrics of Precision and Recall, we define the coverage metrics as:

\begin{align}\small
    \text{COV-R (Recall)} &= \frac{1}{L} | \{ l \in [1..L]: \exists k \in [1..K], D_{\mathrm{struc.}}(\mM_k, \mM_l^*) < \delta_{\mathrm{struc.}}, \nonumber \\
    & \qquad \qquad \qquad \qquad \qquad \qquad \quad D_{\mathrm{comp.}}(\mM_k, \mM_l^*) < \delta_{\mathrm{comp.}} \} | \\
    \text{AMSD-R (Recall)} &= \frac{1}{L} \sum_{l \in [1..L]} \min_{k \in [1..K]} D_{\mathrm{struc.}}(\mM_k, \mM_l^*) \\
    \text{AMCD-R (Recall)} &= \frac{1}{L} \sum_{l \in [1..L]} \min_{k \in [1..K]} D_{\mathrm{comp.}}(\mM_k, \mM_l^*),
\end{align}
where COV is "Coverage", AMSD is "Average Minimum Structure Distance", AMCD is "Average Minimum Composition Distance", and COV-P (precision), AMSD-P (precision), AMCD-P (precision) are defined as in above equations, but with the generated and ground truth material sets swapped. The recall metrics measure how many ground truth materials are correctly predicted, while the precision metrics measure how many generated materials are of high quality (more discussions can be found in \cite{ganea2021geomol}).

We note several points on why we define the metrics in their current forms. 1) COV requires \textit{both} structure and composition distances to be within the thresholds, because generating materials that are structurally close to one ground truth material and compositionally close to another is not meaningful. As a result, AMSD and AMCD are less useful than COV. 2) We use fingerprint distance, rather than RMSE from \texttt{StructureMatcher} \citep{ong2013python}, because the material space is too large for the models to generate enough materials to \textit{exactly} match the ground truth materials. \texttt{StructureMatcher} first requires the compositions of two materials to exactly match, which will cause all models to have close-to-zero coverage.

\begin{figure}[htp]
\REVISION{
    \centering
    \includegraphics[width=0.6\textwidth]{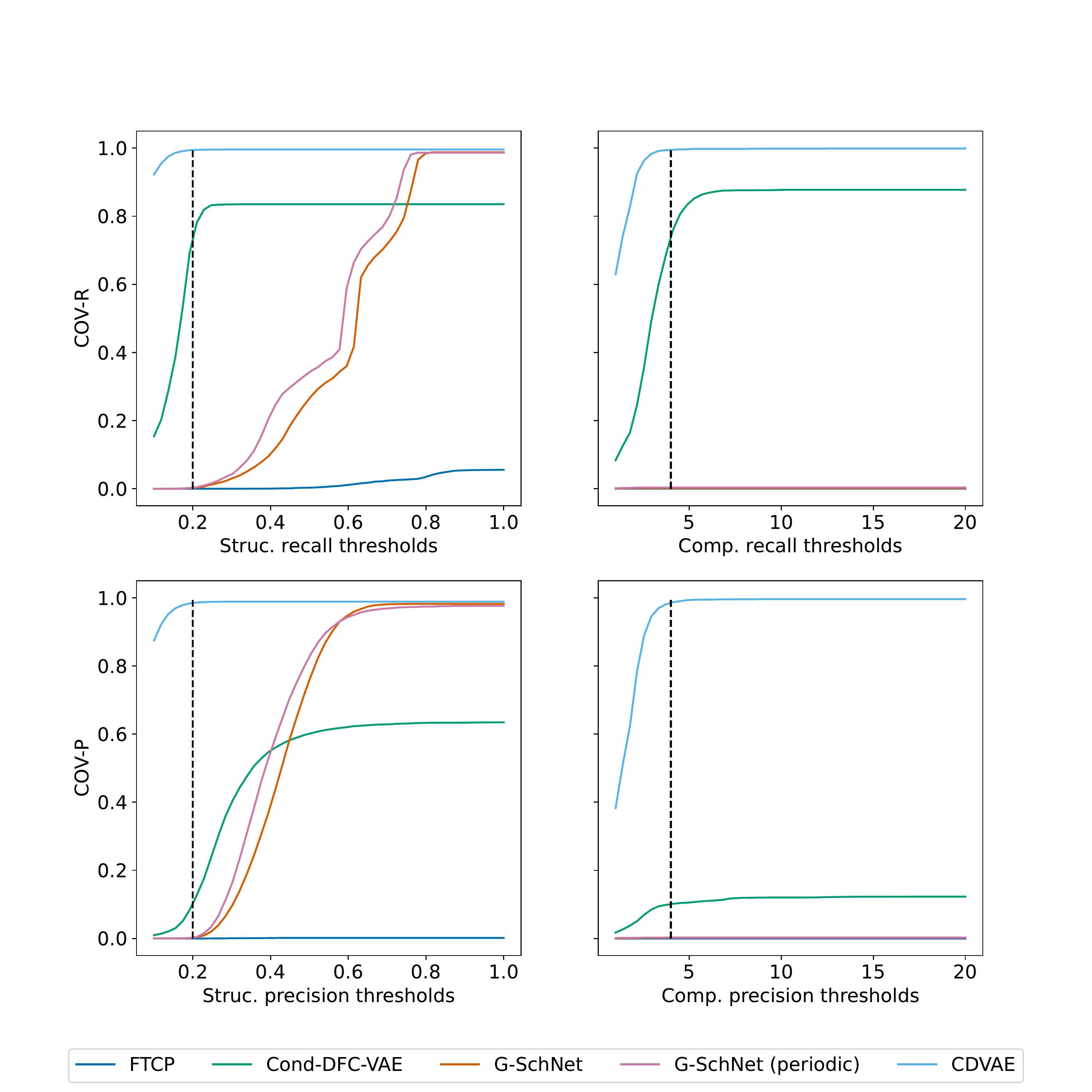}
    \caption{Change of COV-R and COV-P by varying $\delta_{\mathrm{struc.}}$ and $\delta_{\mathrm{comp.}}$ for Perov-5. Dashed line denotes the current chosen thresholds.}
    \label{fig:perov_cov}
}
\end{figure}

\begin{figure}[htp]
\REVISION{
    \centering
    \includegraphics[width=0.6\textwidth]{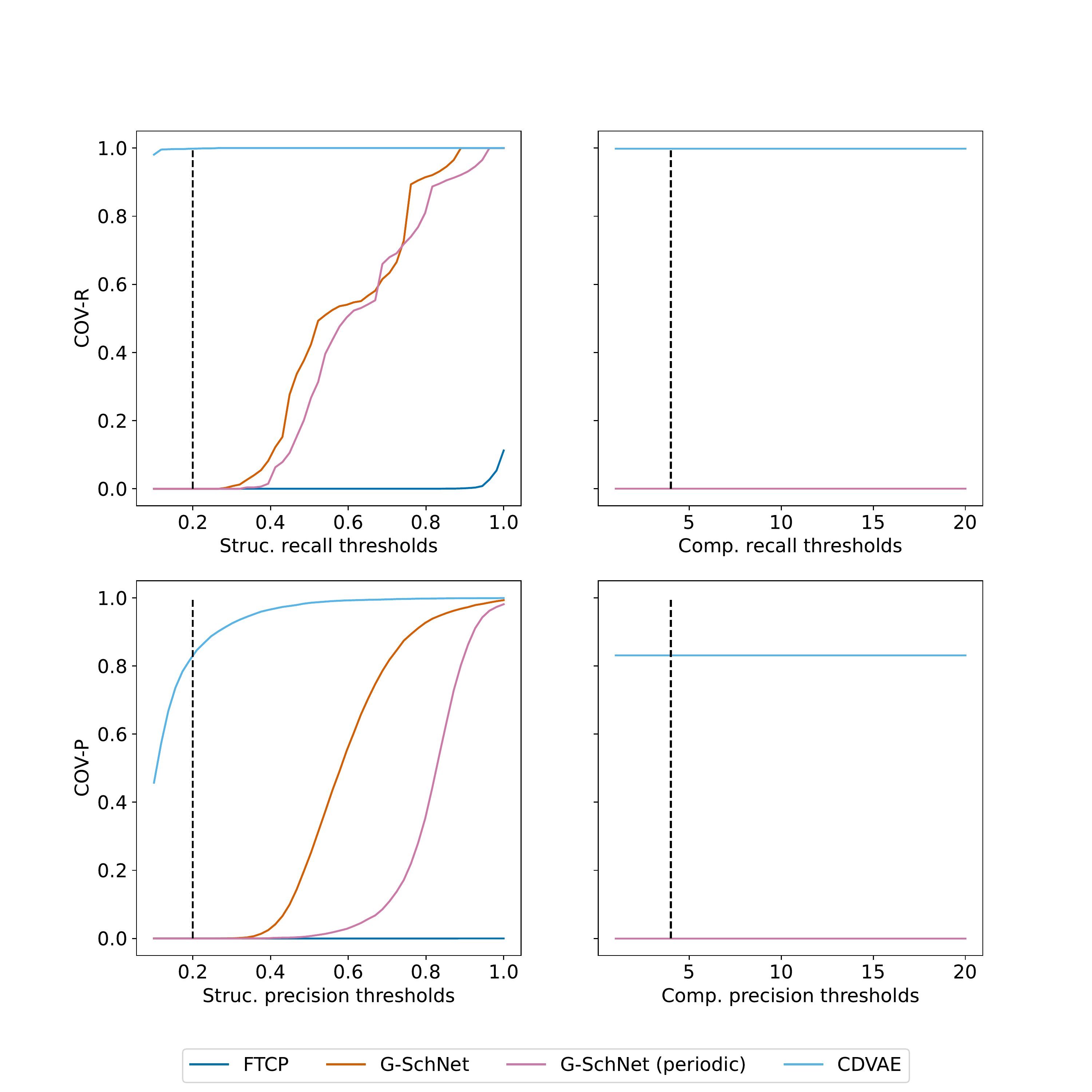}
    \caption{Change of COV-R and COV-P by varying $\delta_{\mathrm{struc.}}$ and $\delta_{\mathrm{comp.}}$ for Carbon-24. Dashed line denotes the current chosen thresholds.}
    \label{fig:carbon_cov}
}
\end{figure}

\begin{figure}[htp]
\REVISION{
    \centering
    \includegraphics[width=0.6\textwidth]{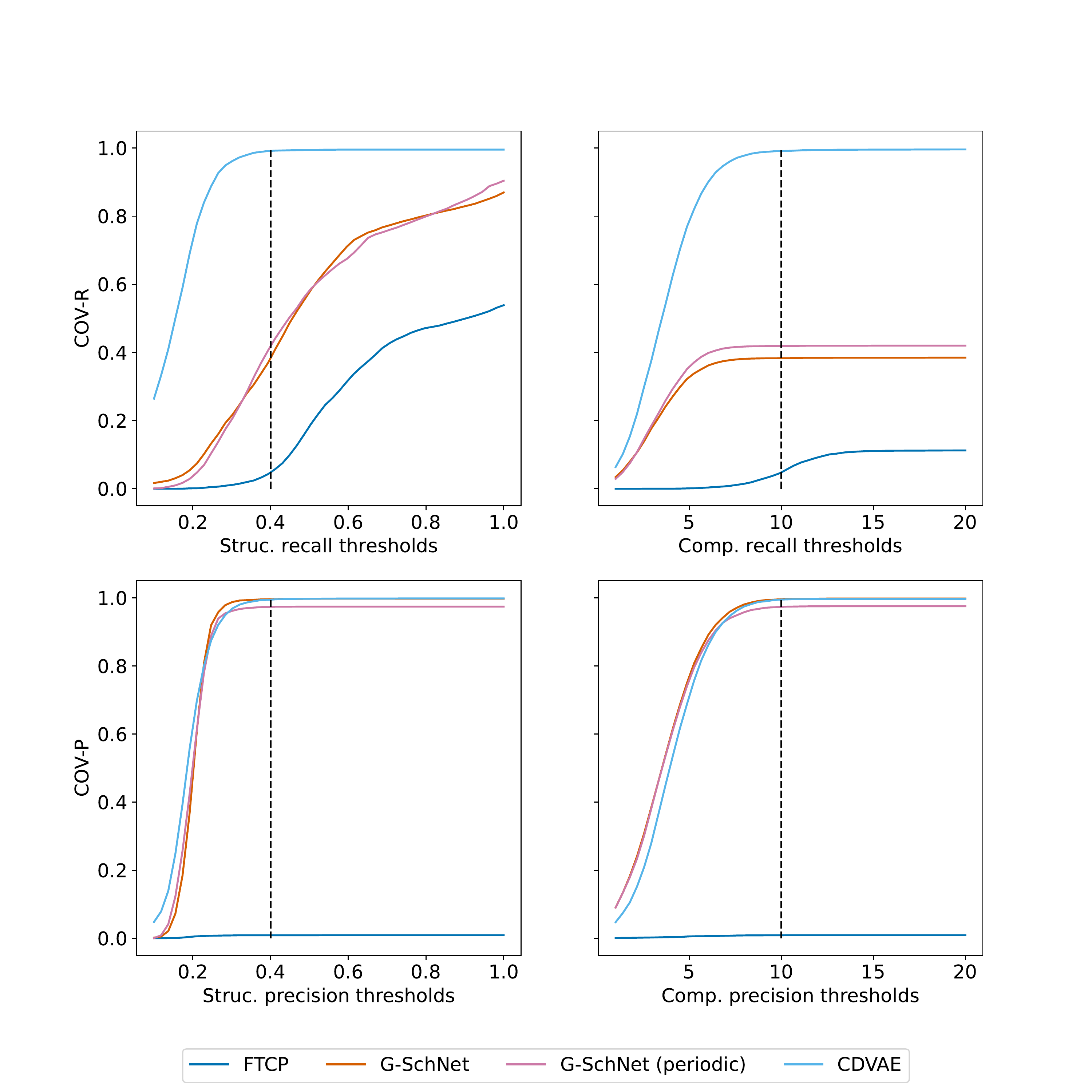}
    \caption{Change of COV-R and COV-P by varying $\delta_{\mathrm{struc.}}$ and $\delta_{\mathrm{comp.}}$ for MP-20. Dashed line denotes the current chosen thresholds.}
    \label{fig:mp_cov}
}
\end{figure}

For Perov-5 and Carbon-24, we choose $\delta_{\mathrm{struc.}} = 0.2, \delta_{\mathrm{comp.}} = 4$. For MP-20, we choose $\delta_{\mathrm{struc.}} = 0.4, \delta_{\mathrm{comp.}} = 10$. In \autoref{fig:perov_cov}, \autoref{fig:carbon_cov}, \autoref{fig:mp_cov}, we show how both COV-R and COV-P change by varying $\delta_{\mathrm{struc.}}$ and $\delta_{\mathrm{comp.}}$ in all three datasets.

\begin{table}[htp]
\REVISION{
\caption{Full coverage metrics for the generation task.}
\label{table:coverage}
\scriptsize
\begin{center}
\begin{tabular}{ll|llllll}
\textbf{Method} & \textbf{Data} & \textbf{COV-R} $\uparrow$ & \textbf{AMSD-R} $\downarrow$ & \textbf{AMCD-R} $\downarrow$ & \textbf{COV-P} $\uparrow$ & \textbf{AMSD-P} $\downarrow$ & \textbf{AMCD-P} $\downarrow$  \\
\hline
\multirow[t]{3}{*}{FTCP} & Perov-5 & 0.00 &	0.7447 &	7.212 &	0.00 &	0.3582 &	3.390 \\ 
& Carbon-24 & 0.00 &	1.181 &	0.00 &	0.00 &	0.8822 &	24.16 \\
& MP-20 & 4.72 &	0.6542 &	9.271 &	0.09 &	0.1954 &	4.378\\

\multirow[t]{1}{*}{Cond-DFC-VAE} & Perov-5 & 73.92 &	0.1508 &	2.773 &	10.13 &	0.3162 &	4.257 \\
\multirow[t]{3}{*}{G-SchNet} & Perov-5 & 0.18 &	0.5962 &	1.006 &	0.23 &	0.4259 &	1.3163 \\
& Carbon-24  & 0.00 &	0.5887 &	0.00 &	0.00 &	0.5970 &	0.00 \\
& MP-20 & 38.33 &	0.5365 &	\textbf{3.233} &	\textbf{99.57} &	0.2026 &	3.601\\
\multirow[t]{3}{*}{P-G-SchNet} & Perov-5 & 0.37 &	0.5510 &	1.0264 &	0.25 &	0.3967 &	1.316  \\
& Carbon-24  & 0.00 &	0.6308 &	0.00 &	0.00 &	0.8166 &	0.00\\
& MP-20 & 41.93 &	0.5327 &	3.274 &	\textbf{99.74} &	0.1985 &	\textbf{3.567} \\
\multirow[t]{3}{*}{CDVAE} & Perov-5 & \textbf{99.45} &	\textbf{0.0482} &	\textbf{0.6969} &	\textbf{98.46} &	\textbf{0.0593} &	\textbf{1.272}\\
& Carbon-24 & \textbf{99.80} &	\textbf{0.0489} &	0.00 &	\textbf{83.08} &	\textbf{0.1343} & 	0.00\\
& MP-20 & \textbf{99.15} &	\textbf{0.1549} &	3.621 &	\textbf{99.49} &	\textbf{0.1883} &	4.014 \\
\end{tabular}
\end{center}
}
\end{table}

\end{document}